\documentclass[twoside]{article}

\usepackage[preprint]{aistats2027}
\usepackage[round]{natbib}
\renewcommand{\bibname}{References}
\renewcommand{\bibsection}{\subsubsection*{\bibname}}

\usepackage{hyperref}
\usepackage{url}
\usepackage{graphicx}
\usepackage{booktabs}
\usepackage{pifont}
\usepackage{algorithmic}
\usepackage{xcolor}
\usepackage{amsmath}
\usepackage{amsfonts}
\usepackage{newfloat}
\usepackage{listings}
\usepackage{subcaption}
\usepackage[most]{tcolorbox}
\usepackage{enumitem}
\usepackage{multirow}
\usepackage{multicol}
\usepackage{wrapfig}
\usepackage{float}

\newcommand{\yesmark}{\ding{51}}
\newcommand{\nomark}{\textcolor{red}{\ding{55}}}\usepackage{algorithm}

\newtcolorbox{promptbox}[1]{
    colback=gray!5,
    colframe=black!60,
    boxrule=0.5pt,
    arc=2pt,
    left=6pt,
    right=6pt,
    top=6pt,
    bottom=6pt,
    title=#1,
    fonttitle=\bfseries,
    breakable
}

\newcommand{\dermseven}{Derm7pt}
\newcommand{\cub}{CUB}
\newcommand{\sun}{SUN}

\newcommand{\qwenllm}{\texttt{Qwen3-32B}}
\newcommand{\llamallm}{\texttt{Llama-3.1-8B-Instruct}}
\newcommand{\phillm}{\texttt{Phi-4-14B}}

\newcommand{\qwenvl}{\texttt{Qwen2-VL-7B-Instruct}}
\newcommand{\internvl}{\texttt{InternVL3-8B}}

\newcommand{\resnet}{\texttt{ResNet-18}}
\newcommand{\blip}{\texttt{BLIP}}

\newcommand{\imageonly}{Image-only}
\newcommand{\aux}{Aux}
\newcommand{\auxunc}{Aux+Unc}
\newcommand{\goldbinary}{Gold Binary}
\newcommand{\goldoriginal}{Gold Original}

\newcommand{\imageaux}{Image+Aux}
\newcommand{\imageauxunc}{Image+Aux+Unc}

\begin{document}

%

%

\twocolumn[

\aistatstitle{Tell Me How Sure You Are: Communicating Uncertainty in Auxiliary Evidence to Foundation Models}

\aistatsauthor{ Soumyajit Chatterjee \And Baturalp Buyukates \And  Abhirup Ghosh }

\aistatsaddress{ Brave \& University of Birmingham\\United Kingdom \And  University of Birmingham\\United Kingdom \And University of Birmingham\\United Kingdom}

]


\begin{abstract}

In many queries, the context information provided to generative foundation models, such as LLMs and VLMs, has inherent uncertainty, including representations of the physical world. While conventional prompting discards the associated uncertainty, this paper, for the first time in the literature, systematically demonstrates that communicating such uncertainty in context provides useful additional cues for downstream decisions and yields improvements across a wide range of settings.

We formulate the decision-making problem as a zero-shot foundation model-assisted image classification task supported by auxiliary text-based characteristics. We choose this due to the availability of well-defined auxiliary properties, associated uncertainty, and unambiguous task correctness. Using five downstream foundation models across three datasets, we extensively evaluate uncertainty arising from both humans and learned models when observing or predicting auxiliary evidence. Across all settings, we find that communicating uncertainty substantially improves downstream LLM classification performance: macro-F1 improves up to $26.18$ percentage points compared to when the uncertainty information is not communicated.

We further observe how the foundation models process such uncertainty information and find that auxiliary uncertainty contains predictive information about the correctness of downstream decisions. Together, these results establish uncertainty as useful contextual information and motivate more systematic consideration of uncertainty communication in future foundation-model inference pipelines.

\end{abstract}

\section{Introduction}
\label{sec:intro}
\begin{figure*}
    \centering
    \includegraphics[width=\textwidth]{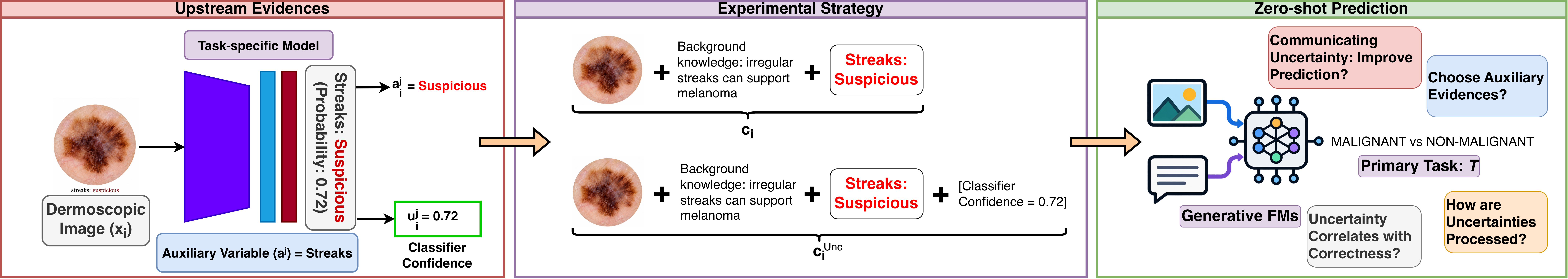}
    \caption{
    Overview of our evaluation framework. Upstream sources provide auxiliary evidence and associated uncertainty. We compare paired zero-shot contexts while holding the auxiliary evidence fixed and varying whether its uncertainty is communicated, and study the resulting effects on downstream prediction, evidence use, and decision correctness.}
    \label{fig:method}
\end{figure*}
Generative foundation models are useful as general-purpose decision layers to combine heterogeneous information at inference time, especially in \emph{zero-shot} settings. Examples include a medical AI assistant aggregating information from clinical records, medical images, and outputs from specialised predictive models processing clinical test results~\citep{liu2023large,moor2023foundation}; or an in-car video agent integrating outputs from specialised object detection models~\citep{zhi2025videoagent}.
In such settings, the advantage of a foundation model is its ability to compose different forms of evidence dynamically, without requiring a new task-specific model for every combination of available information. Nevertheless, these pieces of auxiliary evidence are not necessarily equally reliable.They may originate from human observations, retrieval systems, sensors, classifiers, or other predictive models, and may therefore be incomplete, ambiguous, or incorrect. Moreover, many such sources naturally expose confidence scores or other native estimates of uncertainty, raising a basic question: \emph{should a downstream foundation model receive only the evidence, or also how certain its source was about that evidence?}


Initial studies suggest that communicating or exploiting uncertainty of the auxiliary evidence can benefit foundation-model inference. For example, prior work has examined how LLMs respond to uncertain external knowledge~\citep{pmlr-v318-ghanbari26a, huang2025trust,nafar2025reasoning}, while SRICE~\citep{zhi2025seeing} and VideoAgent2~\citep{zhi2025videoagent} incorporate tool confidence into multimodal agentic pipelines. These results provide promising evidence that uncertainty can matter. However, they do not systematically isolate the effect of \emph{communicating uncertainty itself}: the underlying evidence is not held fixed across conditions, uncertainty may influence subsequent retrieval or routing, and the effect is not examined across different sources of uncalibrated uncertainty. Here we fill all these gaps.

We not only require the auxiliary evidence to be accompanied with \emph{explicit uncertainty or confidence signal} supplied by its source, but also the following: Downstream task \emph{correctness can be measured objectively}, and the \emph{same auxiliary evidence} can be communicated to the foundation model either with or without its additional reliability information. This enables a controlled paired comparison in which the evidence itself is fixed and only the communicated uncertainty changes.


We therefore study three benchmarks that satisfy these requirements while covering distinct evidence-generation processes. In \cub{}~\citep{wah2011caltech}, human annotators provide visual characteristics of bird images together with ordinal observational certainty labels; in \sun{}~\citep{patterson2012sun}, a general-purpose vision model provides image--attribute matching scores; and in \dermseven{}~\citep{Kawahara2018-7pt}, specially trained task-specific classifiers provide dermatological findings together with class probability based confidence. We intentionally retain these signals in the form provided by each source rather than forcing them into a common universal numerical representation, since our aim is not to compare their scales directly but to ask whether communicating the reliability information available at the source helps the downstream model. Using both LLMs and VLMs, we examine their effects on downstream classification performance, how those effects change with the amount and composition of auxiliary evidence, and whether upstream uncertainty is informative about downstream zero-shot decision correctness and the language model's own uncertainty.

Our results reveal four main patterns. First, explicitly communicating uncertainty substantially improves downstream correctness. Across the LLMs, the largest gains reach $7.84$ percentage points (pp) Macro F1 on \cub{}, $26.18$ pp on \sun{}, and $9.10$ pp on \dermseven{}, with smaller but measurable gains also observed for VLMs. Second, we show that uncertainty helps to sanitize a mixed group of certain and uncertain evidence. Third, we observe that the foundation model doesn't merely consider the ordering of evidence in the prompt but inspects the uncertainty values to weight the evidence correctly.
And finally, upstream uncertainty can predict downstream correctness, but does not correlate with the downstream model's own predictive uncertainty: communicating uncertainty can correct a decision without reducing downstream entropy.



Together, these results show that uncertainty is useful contextual information, but its effect depends on the evidence it accompanies and how the downstream model processes that evidence. Our contributions are:

\begin{enumerate}
    \item We introduce a paired zero-shot evaluation framework that isolates the effect of communicating uncertainty while holding the downstream model, primary context, and auxiliary evidence fixed. Across settings the gain in Macro F1 improvement is substantial.

    \item We demonstrate this effect across three distinct evidence-generation processes -- human judgments, a general-purpose vision model, and specialized classifiers 

    \item We show that the benefit depends on if the in-context uncertainty is associated with the evidence: on \sun{}, correctly aligned uncertainty improves macro-F1 by $11.71$--$26.60$ pp relative to omitting uncertainty and by $14.02$--$19.53$ pp relative to misaligned uncertainty. We further show that upstream uncertainty is informative about downstream correctness, yet not directly reflected in the downstream model's own predictive uncertainty.
\end{enumerate}
\section{Literature Survey}
\label{sec:rel_work}
We briefly review prior work on uncertain auxiliary evidence, intermediate representations, and upstream--downstream uncertainty; an expanded discussion appears in Appendix~\ref{app:broad_rel_work}.

\begin{table*}[t]
\centering
\caption{Comparison with relevant prior works. A check denotes that the corresponding criterion is explicitly evaluated. ``Evidence held fixed'' means that auxiliary evidence is unchanged and only the communicated uncertainty varies. ``Distinct uncertainty source types'' refers to qualitatively different signals, such as human-reported certainty, model matching scores, and classifier confidence.}
\label{tab:closest_uncertainty_work}
\resizebox{\textwidth}{!}{
\setlength{\tabcolsep}{4.5pt}
\renewcommand{\arraystretch}{1.18}

\begin{tabular}{lcccccc}
\toprule
& \multicolumn{3}{c}{\textbf{Uncertainty Communication}}
& \multicolumn{2}{c}{\textbf{Downstream Analysis}}
& \textbf{Scope} \\
\cmidrule(lr){2-4}
\cmidrule(lr){5-6}
\cmidrule(l){7-7}

\textbf{Method}
& \shortstack{\textbf{Communicates}\\\textbf{Uncertainty}}
& \shortstack{\textbf{Holds Evidence}\\\textbf{Fixed}}
& \shortstack{\textbf{Measures Effect of}\\\textbf{Uncertainty Communication}}
& \shortstack{\textbf{Analyzes Downstream}\\\textbf{Correctness}}
& \shortstack{\textbf{Analyzes Downstream}\\\textbf{Uncertainty}}
& \shortstack{\textbf{Studies Distinct}\\\textbf{Uncertainty Source Types}} \\
\midrule

SRICE~\citeyearpar{zhi2025seeing}
    & \yesmark & \nomark & \yesmark & \nomark & \nomark & \nomark \\

VideoAgent2~\citeyearpar{zhi2025videoagent}
    & \yesmark & \nomark & \yesmark & \nomark & \nomark & \nomark \\

Tools in the Loop~\citeyearpar{tools-uncertainty,lymperopoulos2025toolsloopquantifyinguncertainty}
    & \nomark & \nomark & \nomark & \yesmark & \nomark & \nomark \\

ProbCBM~\citeyearpar{pmlr-v202-kim23g}
    & \nomark & \nomark & \nomark & \yesmark & \yesmark & \nomark \\

\midrule
\textbf{Our study}
    & \yesmark & \yesmark & \yesmark & \yesmark & \yesmark & \yesmark \\
\bottomrule
\end{tabular}
}
\end{table*}

Prior work has studied uncertain external information through probabilistic reasoning~\citep{nafar2025reasoning}, knowledge revision~\citep{pmlr-v318-ghanbari26a}, calibrated routing~\citep{zhi2025seeing}, and agentic retrieval ~\citep{zhi2025videoagent,huang2025trust,wu2024pandora}. 
Foundation models have been used previously to build interpretable image classifiers through 
concept-based methods using attributes, probabilistic concepts, uncertainty-guided interventions, and descriptor-based zero-shot recognition~\citep{pmlr-v119-koh20a,pmlr-v202-kim23g,park2026analysis,menon2023visual,CBM2026}. 
Foundation-model uncertainty is commonly estimated through self-evaluation, entropy, verbalized confidence, calibration, or tool-level uncertainty~\citep{kadavath2022language,kuhn2023semantic,xiong2024can,chen2026query,yadkori2024believe,tools-uncertainty,lymperopoulos2025toolsloopquantifyinguncertainty}. 

\textbf{Gaps and novelty.} \tablename~\ref{tab:closest_uncertainty_work} shows that prior work addresses complementary parts of this problem.
In summary, no prior work isolates the marginal effect of uncertainty communication by holding the auxiliary evidence fixed and examining its impact on final predictions and correctness. None considered downstream predictive uncertainty across heterogeneous sources of uncertainty. To the best of our knowledge, this is the first study to provide this joint evaluation across evidence sources, domains, and foundation models, as summarized in \figurename~\ref{fig:method}.
\section{Experimental strategy}
\label{subsec:experimental_setup}
\paragraph{Problem formulation.} 
Let a classification dataset, $D$ contain samples of the form $\{\mathbf{x}_i, \mathbf{a}_i, y_i\}$ where, $\mathbf{x}_i\in \mathcal{X}$ is an image and $y_i\in \mathcal{Y}$ is the classification category. Let us call this classification task the {\em primary task}, $\tau$. Auxiliary variables, $\mathbf{a}_i$ capture a set of categorical characteristics, $\mathcal{A}$ of the image $\mathbf{x}_i$. An auxiliary attribute $j$, $a_i^j \in \mathcal{L}^j$ denotes its observed value for sample $i$, where $\mathcal{L}^j$ is the set of possible values for that attribute. The observed value $a_i^j$, therefore, constitutes auxiliary evidence that can provide a cue for predicting the primary-task label $y_i$. For a bird classification task, the color of the wings is an auxiliary variable, and its value for a particular sample bird will be the evidence. {\em Auxiliary tasks} identify the evidence of the auxiliary variables.

We assume that both the image, $\mathbf{x}_i$, and the auxiliary variables $\mathbf{a}_i$ independently have enough information to build a reasonably accurate classifier for $\mathcal{Y}$. This follows from the literature that used such evidence to build interpretable classifiers~\citep{pmlr-v119-koh20a}.

While one can train an end-to-end image classifier $f:\mathcal{X}\rightarrow\mathcal{Y}$ or an auxiliary classifier $g$ that predicts $y_i$ from the auxiliary evidence $\mathbf{a}_i$, our goal in this study is not to build the most accurate end-to-end classifier. Instead, we use a generative foundation model, $\theta$, to predict the label $y_i$ from context information $c_i$ that includes the auxiliary evidence, in a \emph{zero-shot} setting.


\paragraph{Recording auxiliary evidence.} Depending on the applications, there can be two types of auxiliary evidence: $i)$ annotation by a human, and $ii)$ predicted by a learned model. Now, for the latter type, one can predict the auxiliary evidence either by $(ii.a)$ training and inferring from a task-specific model; or $(ii.b)$ inferring from a pre-trained general-purpose vision language model. The choice often depends on whether there is enough data to train an auxiliary evidence predictor or whether the auxiliary task is sufficiently general and suitable for a general-purpose pre-trained model. In this paper, we select datasets that cover all these cases.

\paragraph{Constructing the context.}
Irrespective of the mechanism for recording auxiliary evidence, we construct the prompt $c_i$ with three sections. The first section specifies the context of the primary task ($\tau$), while the second section defines the auxiliary variables ($\mathcal{A}$) and their relations to $\tau$. The final section specifies the auxiliary evidence. Thus, let us represent the context, $c_i$ as a function, $(\tau, \mathcal{A}, \mathbf{a}_i)$. Examples and more details on context construction are in Appendix~\ref{app:prompt_templates}.

While it is well known that optimizing prompts can improve performance, we do not do so in this paper for two reasons. First, when comparing performance across multiple foundation models, optimizing prompts is an open problem itself~\citep{zhuo2024prosa}. Second, results on a variety of prompt configurations (Appendix~\ref{app:derm_prompt_sensitivity}) indicate that the observed benefit is generally robust to prompt wording, although its magnitude varies across prompt formulations.

\paragraph{Uncertainty of auxiliary variables.} In the context of the applications considered, auxiliary variables $\mathbf{a}_i$ are observed with uncertainty $\mathbf{u}_i$, i.e., each evidence, $a_i^j$, is associated with an uncertainty value, $u_i^j$. 
Throughout, we use \emph{uncertainty signal} as an umbrella term for source-provided reliability information; depending on the source, this may be expressed as uncertainty, certainty, or confidence.

The type of uncertainty directly depends on how the auxiliary evidence is recorded in an application. When a human annotates an auxiliary variable $(i)$, the uncertainty originates from either a lack of precision in defining the variable or visual noise in the sample. Note that the uncertainty signals obtained from $(ii.a)$ a task-specific model and $(ii.b)$ a general-purpose foundation model can have different semantics and scales, and we do not assume that they are calibrated or directly comparable. There are various ways to measure uncertainty in task-specific models, such as the entropy of the class probabilities or the probability of the most likely class \citep{KangUnc}. While Table~\ref{tab:derm_prompt_sensitivity} evaluates several uncertainty representations, including entropy, maximum softmax confidence, and their combination, across multiple prompt configurations for the task-specific model, in the rest of the experiments, we use probability of the most likely auxiliary class and BLIP image--text matching scores as the uncertainty signals for the task-specific model and general-purpose foundation model, respectively. 

We assume that all uncertainty values within a context share the same representation and interpretation; mixing heterogeneous uncertainty representations within a single context is outside the scope of this work. To communicate uncertainty to the foundation model, we keep the same prompt structure as above and append the uncertainty information in the final section along with the auxiliary evidence. Thus the context, $c_i^{\texttt{Unc}}$ becomes $(\tau, \mathcal{A}, \mathbf{a}_i, \mathbf{u}_i)$.

\subsection{Datasets}
We use three datasets exhaustively covering the scenarios ($i$, $ii.a$, $ii.b$) discussed above.

\noindent\textbf{\cub{}.}
The primary task in \cub{}~\citep{wah2011caltech} is to classify bird species from images. While the dataset contains $200$ species and $312$ human-annotated visual attributes, to fit the context length we consider candidate-species configurations containing $7$, $8$, and $9$ classes; the class-selection procedure and additional details are provided in Appendix~\ref{app:cub_classes}. For each sample $i$, $\mathbf{a}_i$ contains only the attributes annotated as present. Each retained evidence $a_i^j$ is accompanied by the ordinal human-certainty signal $u_i^j\in\{\textit{guessing},\textit{probably},\textit{definitely}\}$. These labels represent the annotator's certainty.

\noindent\textbf{\sun{}.}
The \sun{} dataset~\citep{patterson2012sun} asks for scene classification and provides a list of visual and contextual scene attributes. We consider four scene categories encompassing common places in a city life -- \textit{park}, \textit{office cubicles}, \textit{supermarket}, and \textit{coffee shop} -- with $20$ images per class, and construct a pool of $20$ auxiliary variables (Appendix~\ref{app:sun_construction} has the details). For each sample $i$, we use the image--text matching head of the pre-trained general-purpose vision--language model \blip{}~\citep{li2022blip} to compute an image--text matching score for each attribute. We rank the $20$ attributes by their \blip{} scores and use the top $K\in\{3,5,8,10\}$ attribute names as the image-specific auxiliary evidence $\mathbf{a}_i$. Each retained evidence $a_i^j$ is associated with the corresponding \blip{} image--text matching score $u_i^j$, which we use as its uncertainty. 

\noindent\textbf{\dermseven{}.}
The \dermseven{} dataset~\citep{Kawahara2018-7pt} asks for skin-lesion classification (primary task, $\tau$) and provides seven dermatological findings (auxiliary variables) from the seven-point checklist. We formulate $\tau$ as a binary classification problem, treating melanoma and basal cell carcinoma as malignant and the remaining diagnoses as non-malignant. For each auxiliary task, we independently train an task-specific \resnet{} classifier~\citep{he2016deep} and map its prediction to a binary state, \textit{not suspicious} or \textit{suspicious} (Appendix~\ref{app:derm7pt_setup}). For each sample $i$, these predicted findings constitute the auxiliary evidence $\mathbf{a}_i$, where each $a_i^j$ is accompanied by the maximum-softmax confidence $u_i^j$ from the upstream classifier.
\begin{table*}[t]
\scriptsize
\centering
\caption{Macro-F1 across \cub{}, \sun{}, and \dermseven{}. $\Delta$ denotes the \auxunc{} gain over \aux{} in percentage points. All experiments use three seeds (42, 96, 1001); \cub{} reports mean $\pm$ standard deviation across seed-specific sampled evaluation sets, while \sun{} and \dermseven{} use fixed evaluation sets and omit $\pm 0.0$ when deterministic inference yields identical results.}
\label{tab:main_results}
\begin{tabular}{llccr}
\toprule
\textbf{Dataset} &
\textbf{Model} &
\textbf{Aux} &
\textbf{Aux + Uncertainty} &
$\boldsymbol{\Delta}$ \textbf{(pp)} \\
\midrule

\multirow{3}{*}{CUB (7-class)}
& Qwen3-32B
& $0.4682 \pm 0.0165$
& $0.5424 \pm 0.0131$
& $\mathbf{+7.42 \pm 0.76}$ \\

& Llama-3.1-8B
& $0.5770 \pm 0.0095$
& $0.6368 \pm 0.0083$
& $+5.98 \pm 0.94$ \\

& Phi-4-14B
& $0.6924 \pm 0.0174$
& $0.7385 \pm 0.0194$
& $+4.61 \pm 0.84$ \\

\midrule

\multirow{3}{*}{CUB (8-class)}
& Qwen3-32B
& $0.3602 \pm 0.0039$
& $0.4387 \pm 0.0156$
& $\mathbf{+7.84 \pm 1.72}$ \\

& Llama-3.1-8B
& $0.5260 \pm 0.0041$
& $0.5753 \pm 0.0109$
& $+4.93 \pm 1.10$ \\

& Phi-4-14B
& $0.5827 \pm 0.0154$
& $0.6083 \pm 0.0051$
& $+2.56 \pm 1.06$ \\

\midrule

\multirow{3}{*}{CUB (9-class)}
& Qwen3-32B
& $0.3351 \pm 0.0076$
& $0.3805 \pm 0.0130$
& $\mathbf{+4.53 \pm 0.61}$ \\

& Llama-3.1-8B
& $0.4920 \pm 0.0070$
& $0.5115 \pm 0.0067$
& $+1.95 \pm 0.22$ \\

& Phi-4-14B
& $0.5638 \pm 0.0102$
& $0.6073 \pm 0.0249$
& $+4.36 \pm 1.50$ \\

\midrule

\multirow{3}{*}{SUN ($K=10$)}
& Qwen3-32B
& 0.6066
& 0.8684
& \textbf{+26.18} \\

& Llama-3.1-8B
& 0.8648
& 0.9241
& +5.93 \\

& Phi-4-14B
& 0.8175
& 0.8843
& +6.69 \\

\midrule

\multirow{3}{*}{Derm7pt ($k=7$)}
& Qwen3-32B
& 0.505
& 0.596
& +9.10 \\

& Llama-3.1-8B
& 0.601
& 0.630
& +2.90 \\

& Phi-4-14B
& 0.544
& 0.547
& +0.25 \\

\bottomrule
\end{tabular}
\end{table*}

\subsection{Generative Foundation Models}
\label{subsec:downstream_models}
We evaluate five general-purpose generative foundation models spanning text-only and multimodal settings. The LLMs are \llamallm{}~\cite{grattafiori2024llama} ($8$B), \phillm{}~\cite{abdin2024phi} ($14$B), and \qwenllm{}~\cite{qwen3technicalreport} ($32$B), covering distinct model families and parameter scales. For VLMs, we use \qwenvl{}~\cite{Qwen2VL} and \internvl{}~\cite{chen2024expanding,wang2024mpo,chen2024far,chen2024internvl}, two similarly sized ($7$--$8$B) models from different families. All models are general-purpose and are evaluated zero-shot without fine-tuning. The LLMs receive only the textual context, while the VLMs additionally receive the image, $\mathbf{x}_i$. Further details are provided in Appendix~\ref{app:downstream_models}.

\begin{table*}[]
\scriptsize
\centering
\caption{CUB VLM macro-F1 across 7-, 8-, and 9-class settings. $\Delta$ denotes the \imageauxunc{} gain over \imageaux{} in percentage points.}
\label{tab:cub_vlm}
\begin{tabular}{lccccc}
\toprule
\textbf{Model} &
\textbf{Classes} &
\textbf{\imageonly{}} &
\textbf{\imageaux{}} &
\textbf{\imageauxunc{}} &
$\boldsymbol{\Delta}$ \\
\midrule

\internvl{}
& 7
& $0.7873 \pm 0.0025$
& $0.7963 \pm 0.0027$
& $0.7993 \pm 0.0036$
& $+0.30$ \\

\qwenvl{}
& 7
& $0.8022 \pm 0.0036$
& $0.8189 \pm 0.0041$
& $0.8183 \pm 0.0042$
& $-0.06$ \\

\midrule

\internvl{}
& 8
& $0.7154 \pm 0.0019$
& $0.7288 \pm 0.0017$
& $0.7313 \pm 0.0055$
& $+0.25$ \\

\qwenvl{}
& 8
& $0.7782 \pm 0.0046$
& $0.7735 \pm 0.0056$
& $0.7942 \pm 0.0067$
& $\mathbf{+2.07}$ \\

\midrule

\internvl{}
& 9
& $0.7370 \pm 0.0069$
& $0.7632 \pm 0.0056$
& $0.7682 \pm 0.0062$
& $+0.50$ \\

\qwenvl{}
& 9
& $0.6862 \pm 0.0011$
& $0.7048 \pm 0.0060$
& $0.7124 \pm 0.0093$
& $+0.76$ \\

\bottomrule
\end{tabular}
\end{table*}

\subsection{Evaluation protocol.}
Our primary metric is macro-F1 for the primary task, and we quantify the effect of uncertainty as the paired change from \aux{} (using context of the form $c_i$) to \auxunc{} (using context of the form $c_i^{\texttt{Unc}}$) in percentage points. All paired comparisons use identical samples and auxiliary evidence, differing only in whether $\mathbf{u}_i$ is communicated. Details of repeated runs and complete results for the remaining evaluated configurations are provided in Appendix~\ref{app:evaluation_protocol} and Appendix~\ref{app:additional_exps}; the main text reports a balanced view across all three benchmarks.

\section{Communicating Uncertainty Improves Downstream Prediction}
\label{sec:finding1}
Here, we test our core hypothesis: whether communicating the uncertainty $u_i^j$ associated with auxiliary evidence $a_i^j$ improves the downstream generative foundation model's final decision. This is important because auxiliary evidence is rarely equally reliable: a human annotator may be unsure about an observed attribute, a vision-language model may weakly match a scene attribute, or a task-specific classifier may assign low confidence to a predicted finding.

\paragraph{Communicating uncertainty consistently improves LLM performance (\tablename~\ref{tab:main_results}).} Across all three benchmarks, communicating $\mathbf{u}_i$ improves downstream LLM performance while keeping the auxiliary evidence fixed. On \cub{}, every reported model--configuration pair improves, with gains of up to $7.84$ percentage points. For \sun{}, all three LLMs improve, with gains ranging from $5.93$ to $26.18$ points; on \dermseven{}, the corresponding gains range from $0.25$ to $9.10$ points. Thus, the core hypothesis holds across human-reported certainty, model scores, and task-specific classifier confidence.

\paragraph{The effect extends to VLMs, although the gains are weaker (\tablename~\ref{tab:cub_vlm}).} On \cub{}, communicating human-reported certainty improves macro-F1 in five of the six VLM settings, with gains reaching $2.07$ percentage points; the only decrease is negligible ($0.06$ points). Thus, uncertainty remains useful even when the downstream model can directly inspect the image, although the effect is smaller than for the text-only LLMs. On \sun{}, both VLMs already achieve macro-F1 of $1.0$, leaving no headroom to measure an additional benefit. The \dermseven{} VLM results show similar modest effects and are reported in 
Appendix~\ref{app:dermseven_vlm_results}.

\section{Selecting Auxiliary variables}
\label{sec:finding2a}
Here, we consider the case where a subset of auxiliary variables ($\mathcal{S} \subset \mathcal{A}$) is given in the context and show that the LLM's accuracy depends on the selection of $\mathcal{S}$. This is important for two reasons. Firstly, where the auxiliary evidence collection is costly, for example, 
in the \dermseven{} dataset, auxiliary evidence corresponds to a clinical test. Secondly, even when 
auxiliary evidence is available, the context window length may be restrictive. For example, the \sun{} dataset contains $102$ auxiliary variables that extend the context beyond the models' context length.

\paragraph{Uncertainty helps when $\mathcal{S}$ includes uncertain evidences (Table~\ref{tab:sun_robustness}).} For each sample, $i$ in \sun{} dataset we rank the auxiliary variables, $a^j_i$ according to their BLIP scores. Then we select $K$ top ranked variables in $\mathcal{S}$ to include in the context ($c_i$ and $c_i^{\texttt{Unc}}$). 

Table~\ref{tab:sun_robustness} reports the number of samples (out of $80$ test samples in this dataset) whose prediction flips from correct to incorrect when context includes more auxiliary evidence. We observe that such flips increase with increasing $K$, making it obvious that unrelated auxiliary variables (with higher $K$) hurt. However, the more interesting fact here is that the number of flips in all cases reduces when we include the uncertainty of the auxiliary evidence, $\mathbf{u}_i$. Further analysis of how evidence behaves depending on its correlation across classes is presented in Appendix~\ref{app:sun_attr_structure}.

\begin{figure}[t]
  \centering
    \centering
    \includegraphics[width=0.84\linewidth, keepaspectratio]{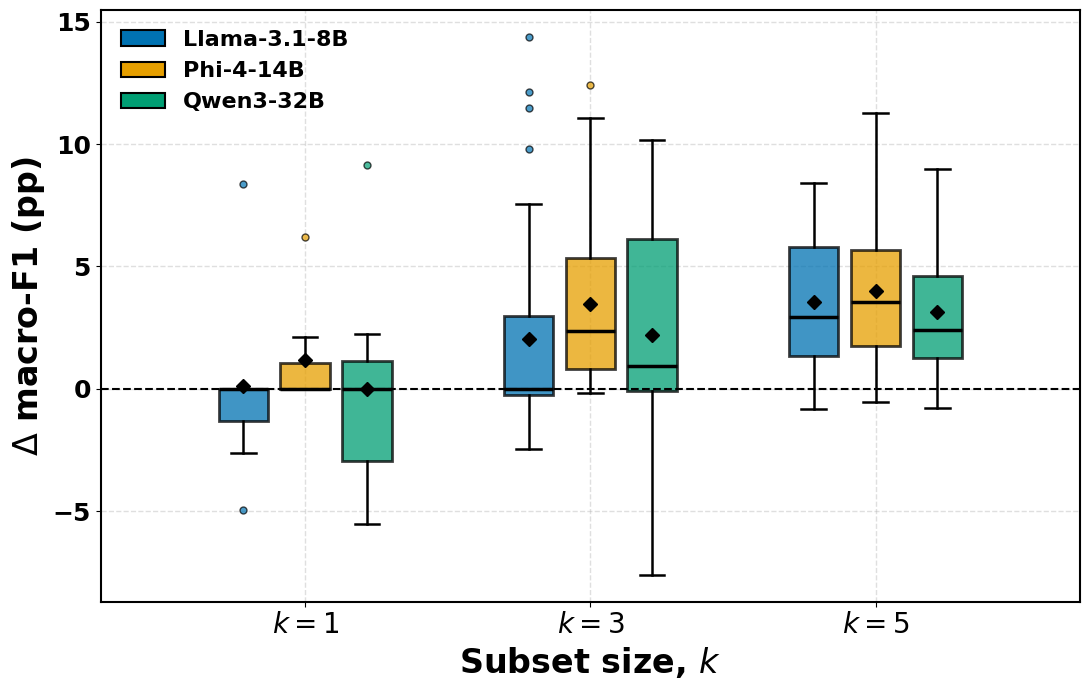}
  \caption{Effect of uncertainty communication across \dermseven{} evidence compositions for LLMs
  Boxes show the distribution of $\Delta$ macro-F1 (pp) across subsets of size $k$; black lines and diamonds denote the median and mean, respectively.
  }
  \label{fig:derm7pt_finding_2a}
\end{figure}

\paragraph{Context should contain important auxiliary information (\figurename~\ref{fig:derm7pt_finding_2a}).} Here, we ask whether the gain from communicating uncertainty changes as more auxiliary variables are included in the context. For \dermseven{}, the gain $\Delta$ generally increases with the number of auxiliary variables for the LLMs. This is consistent with only a subset of auxiliary variables being strongly informative for the primary task $\tau$, while additional variables may be less relevant or less reliable. Communicating their uncertainty can therefore help prevent such evidence from disproportionately affecting the prediction for $\tau$. The VLM results are less uniform, suggesting that this effect also depends on the downstream model.

For completeness, Appendix~\ref{app:subset_results} reports the extended \sun{} and \dermseven{} results across all evaluated auxiliary-variable subsets.

\section{How Uncertainties are Processed?}
\label{sec:finding2b}

Here, we examine how the downstream model uses uncertainty to weight auxiliary variables in a zero-shot way, without task-specific training. We show this in two ways. First, we observe that $\theta$ processes the uncertainty values, not merely looking at the order of the variables in the prompt.
Second, performance is affected substantially by whether important auxiliary variables are present in the context. This is important because not all auxiliary variables contribute equally to the primary task.

\begin{table}[]
\centering
\small
\caption{Number of predictions that are correct at the smaller top-$K$ setting but become incorrect after lower-ranked auxiliary attributes are added on \sun{}. Communicating the corresponding BLIP scores substantially reduces these harmful transitions.
}
\label{tab:sun_robustness}
\resizebox{\columnwidth}{!}{
\begin{tabular}{lcccccc}
\toprule
& \multicolumn{2}{c}{\textbf{$K:5\rightarrow8$}}
& \multicolumn{2}{c}{\textbf{$K:8\rightarrow10$}}
& \multicolumn{2}{c}{\textbf{$K:3\rightarrow10$}} \\
\cmidrule(lr){2-3}
\cmidrule(lr){4-5}
\cmidrule(lr){6-7}
\textbf{Model}
& \textbf{Aux}
& \textbf{Aux+Unc}
& \textbf{Aux}
& \textbf{Aux+Unc}
& \textbf{Aux}
& \textbf{Aux+Unc} \\
\midrule
Qwen3-32B     & 6 & 2 & 7 & 4 & 7 & 2 \\
Llama-3.1-8B  & 5 & 1 & 5 & 0 & 5 & 2 \\
Phi-4-14B     & 2 & 0 & 4 & 2 & 6 & 3 \\
\midrule
\textbf{Total}
& \textbf{13} & \textbf{3}
& \textbf{16} & \textbf{6}
& \textbf{18} & \textbf{7} \\
\bottomrule
\end{tabular}
}
\end{table}

\begin{table*}[]
\centering
\scriptsize
\caption{SUN uncertainty-alignment control at $K=10$. The same ten BLIP-selected attributes are presented in a score-independent random order under all conditions. \textbf{No-U} omits BLIP scores; \textbf{True-U} attaches each attribute's correct score; and \textbf{Shuffled-U} preserves the same score multiset but reassigns scores to different attributes. Values are macro-F1 (mean $\pm$ standard deviation across three random-order seeds); $\Delta$ is reported in percentage points.}
\label{tab:sun_alignment_control}
\begin{tabular}{lccccc}
\toprule
\textbf{Model}
& \textbf{No-Uncertainty}
& \textbf{True-Uncertainty}
& \textbf{Shuffled-Uncertainty}
& $\boldsymbol{\Delta}_{\textbf{True-No}}$ \textbf{(pp)}
& $\boldsymbol{\Delta}_{\textbf{True-Shuffled}}$ \textbf{(pp)} \\
\midrule
\qwenllm{}
& $0.3902 \pm 0.0638$
& $0.6562 \pm 0.0414$
& $0.4609 \pm 0.0105$
& $+26.60 \pm 3.12$
& $+19.53 \pm 3.12$ \\

\llamallm{}
& $0.5630 \pm 0.0270$
& $0.6800 \pm 0.0399$
& $0.5398 \pm 0.0386$
& $+11.71 \pm 5.13$
& $+14.02 \pm 6.08$ \\

\phillm{}
& $0.3846 \pm 0.0304$
& $0.6439 \pm 0.0760$
& $0.4694 \pm 0.0274$
& $+25.93 \pm 8.36$
& $+17.44 \pm 5.03$ \\
\bottomrule
\end{tabular}
\end{table*}
\begin{figure}
    \centering
    \includegraphics[width=\columnwidth, keepaspectratio]{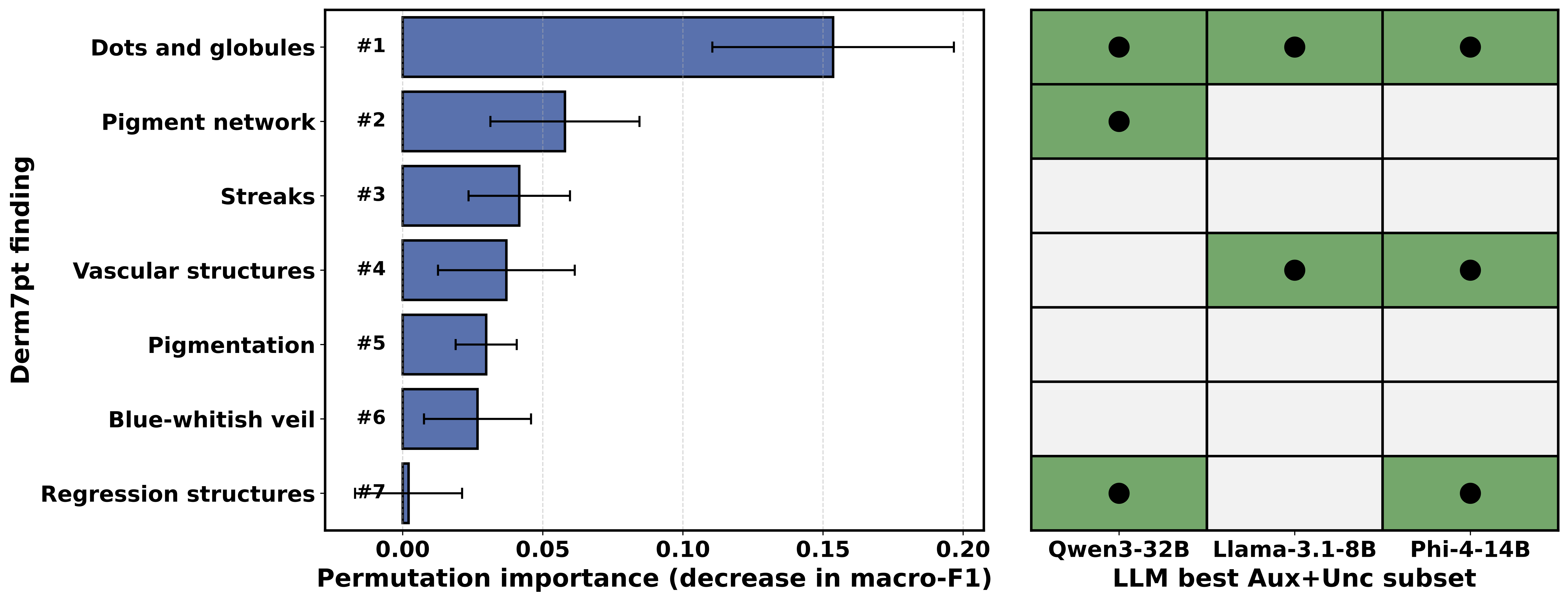} \caption{Supervised feature relevance versus the best-performing \auxunc{} evidence subsets on \dermseven{}. Bars show Random-Forest permutation importance~\citep{breiman2001random}; filled cells indicate findings selected in each LLM's best subset. 
    }
    \label{fig:derm_rf_alignment}
\end{figure}
\begin{table}[t]
\scriptsize
\centering
\caption{Downstream correctness prediction across CUB (from human-reported attribute certainty) and SUN (from four BLIP-derived reliability features, $K=5$, using a stratified five-fold logistic-regression probe). Values report pooled out-of-fold AUROC.}
\label{tab:cub_sun_correctness}
\begin{tabular}{lll}
\toprule
\textbf{Dataset} & \textbf{Model} & \textbf{AUROC} \\
\midrule
\multirow{2}{*}{CUB} 
  & \phillm{} & $0.558 \pm 0.056$ \\
  & \qwenvl{} & $0.472 \pm 0.084$ \\
\midrule
\multirow{2}{*}{SUN} 
  & Qwen3-32B & $0.8002$ \\
  & Phi-4-14B & $0.8150$ \\
\bottomrule
\end{tabular}
\end{table}

\paragraph{Uncertainty is useful only when correctly associated with the evidence (Table~\ref{tab:sun_alignment_control}).} In Table~\ref{tab:sun_alignment_control}, we show that the gain from communicating uncertainty depends on its correct alignment with the auxiliary evidence. Across all three LLMs, correctly aligned uncertainty performs better than both omitting uncertainty and reassigning the same uncertainty values across attributes. To test this, we first randomly order the same selected auxiliary attributes and then compare contexts with no uncertainty, with the correct uncertainty $u_i^j$ associated with each evidence item $a_i^j$, and with a perturbed association in which $a_i^j$ is paired with $u_i^{\pi(j)}$.\footnote{$\pi$ denotes a permutation of the auxiliary-variable indices, so that the same set of uncertainty values is preserved but assigned to different evidences.} We use \sun{} for this experiment because its auxiliary attributes can be ranked by their BLIP-derived matching scores, with an uncertainty value available for each attribute. This allows us to perturb the attribute ordering and the uncertainty--attribute association while keeping the selected evidence fixed.

\begin{table}[t]
\centering
\small
\caption{Predicting \dermseven{} downstream decision correctness from upstream classifier uncertainty at $k=7$. For all models, AUROC is obtained using a standardized logistic-regression probe with the indicated uncertainty features under stratified five-fold cross-validation. We report pooled out-of-fold AUROC.}
\label{tab:derm7pt_correctness}
\resizebox{\columnwidth}{!}{
\begin{tabular}{lllcc}
\toprule
 \textbf{Model} & \textbf{Upstream Features}
& \textbf{AUROC} \\
\midrule
\multirow{3}{*}{\llamallm{}} & Confidence & 0.6259 \\
& Entropy & 0.6257 \\
& Full & 0.6480 \\

\midrule
 \multirow{3}{*}{\qwenvl{}}
& Confidence & 0.6388 \\
& Entropy & 0.6377 \\
& Full & 0.6358 \\

\bottomrule
\end{tabular}
}
\vspace{-1.5em}
\end{table}
\begin{table*}[h]
\centering
\scriptsize
\caption{Changes in downstream predictive entropy on \dermseven{} at $k=7$ after communicating upstream classifier confidence. $\Delta H = H_{\mathrm{With~\texttt{Unc}}}-H_{\mathrm{Without~\texttt{Unc}}}$; negative values indicate reduced predictive uncertainty. The final column reports the percentage of samples with $\Delta H<0$.}
\label{tab:derm7pt_downstream_entropy}
\begin{tabular}{lllrrrr}
\toprule
\textbf{Modality} &
\textbf{Model} &
\textbf{Correctness transition} &
$\boldsymbol{H_{\mathrm{Without~\texttt{Unc}}}}$ &
$\boldsymbol{H_{\mathrm{With~\texttt{Unc}}}}$ &
$\boldsymbol{\Delta H}$ &
$\boldsymbol{\Delta H<0}$ (\%) \\
\midrule

\multirow{4}{*}{LLM}

& \multirow{4}{*}{\llamallm{}}
& Incorrect $\rightarrow$ Correct   & 0.6736 & 0.6861 & +0.0125 & 48.3 \\
& & Correct $\rightarrow$ Incorrect & 0.6836 & 0.6850 & +0.0015 & 52.4 \\
& & Correct $\rightarrow$ Correct   & 0.4498 & 0.4737 & +0.0239 & 34.6 \\
& & Incorrect $\rightarrow$ Incorrect & 0.5883 & 0.6065 & +0.0182 & 39.3 \\

\midrule

\multirow{4}{*}{VLM}
& \multirow{4}{*}{\qwenvl{}}
& Incorrect $\rightarrow$ Correct   & 0.6315 & 0.6773 & +0.0457 & 58.3 \\
& & Correct $\rightarrow$ Incorrect & 0.6885 & 0.6796 & -0.0089 & 40.0 \\
& & Correct $\rightarrow$ Correct   & 0.2406 & 0.2687 & +0.0280 & 39.9 \\
& & Incorrect $\rightarrow$ Incorrect & 0.2412 & 0.3230 & +0.0819 & 18.9 \\

\bottomrule
\end{tabular}
\end{table*}

\paragraph{Useful evidence compositions partially align with supervised feature relevance (\figurename~\ref{fig:derm_rf_alignment}).} For \dermseven{}, we compare the best-performing evidence subset $\mathcal{S}$ for each LLM with an independent Random-Forest feature ranking computed from the training data using permutation importance~\citep{breiman2001random}.
The highest-ranked supervised feature, \textit{dots and globules}, appears in the best subset for all three LLMs, while other highly ranked findings also recur across models. However, the agreement is only partial: some best-performing subsets also contain findings with low supervised importance. Thus, the downstream models do not simply recover the supervised feature ranking; their preferred evidence compositions show signs of implicit \emph{zero-shot feature prioritization}, in which useful auxiliary variables can be identified without task-specific training of the downstream model.

\section{Uncertainty Correlates with Downstream Decisions}
\label{sec:finding3}
Here, we consider whether the uncertainty $\mathbf{u}_i$ associated with the auxiliary evidence is informative about the downstream decision itself. This is important because two identical downstream decisions may be supported by evidence of very different reliability. 
For example, in \dermseven{}, consider a lesion image that receives the same final prediction when the supporting dermatological findings are predicted with either high or low confidence. If the downstream generative foundation model meaningfully processes upstream uncertainty, it should therefore be related to whether the final decision is correct and, potentially, to the model's own predictive uncertainty, which we denote by $H_i$.\footnote{$H_i$ is computed from the downstream model's predictive distribution; the procedure used to obtain class probabilities for LLMs and VLMs is described in Appendix~\ref{app:downstream_uncertainty}.} We examine both relationships below.

\paragraph{Upstream uncertainty is informative of downstream correctness (\tablename~\ref{tab:cub_sun_correctness} \& \tablename~\ref{tab:derm7pt_correctness}).} We examine whether the uncertainty associated with auxiliary evidence can distinguish correct from incorrect downstream decisions; the uncertainty features used for each benchmark are detailed in Appendix~\ref{app:uncertainty_features}. In \cub{}, human-reported certainty is only weakly informative, with AUROC remaining close to chance for both tested models. In contrast, the BLIP-derived reliability profile in \sun{} is strongly predictive of LLM correctness at $K=5$. In \dermseven{}, classifier-derived uncertainty provides a weaker but consistent signal for both the LLM and VLM across the evaluated uncertainty representations. Overall, upstream uncertainty is informative of downstream correctness, although the strength of this relationship depends on the source of uncertainty and the downstream model.


\paragraph{Improved decisions do not necessarily imply lower downstream uncertainty (\tablename~\ref{tab:derm7pt_downstream_entropy}).}
On \dermseven{}, we ask whether communicating upstream classifier confidence is reflected in the downstream model's own predictive uncertainty $H_i$. The two are not directly coupled: even when uncertainty communication changes an incorrect decision to a correct one, downstream entropy often increases rather than decreases, for both the LLM and VLM. Thus, upstream uncertainty can affect the downstream decision without being directly mirrored by the model's own predictive uncertainty. We examine this upstream-to-downstream uncertainty relationship more directly through a Bernoulli view in Appendix~\ref{app:unc_to_unc}.

\section{Conclusions}
\label{sec:conclusion}
\paragraph{Limitations and future scope.} We focus on zero-shot classification, which enables objective measurement of downstream correctness but does not capture open-ended generation or question-answering settings. We do not consider the case where the uncertainty information itself comes from a heterogeneous sources in a single context. 
Extending the analysis to open-ended tasks, cross-source calibration, and uncertainty-aware prompt optimization, therefore, remains a useful direction for future work.

To conclude, we studied whether communicating uncertainty associated with auxiliary evidence improves zero-shot generative foundation-model inference. Across three benchmarks, three LLMs, and two VLMs, communicating uncertainty generally improves downstream classification performance. In the main reported settings, LLM macro-F1 increases by $1.95$--$7.84$ points on \cub{}, $5.93$--$26.18$ on \sun{}, and $0.25$--$9.10$ on \dermseven{}; we observe similar patterns for VLMs, although the gains are generally smaller. The benefit depends on the ordering, alignment, and composition of auxiliary evidence, while upstream uncertainty informs downstream correctness without being directly reflected in the model's own predictive uncertainty. These results support treating uncertainty as part of the context supplied to generative foundation models rather than discarding it before inference.

\clearpage
\subsection*{AI use statement}
We used LLM-based tools to assist with writing and editing the manuscript, and with writing, debugging, and refining research code. All LLM-assisted text was reviewed and edited by the authors, as was all LLM-generated or modified code. LLMs were not relied upon to make any final scientific or experimental decisions. The authors take full responsibility for the final content of this work, including all text, claims, results, code, and other artifacts produced with the aid of LLMs.

\bibliography{reference}

\clearpage
\appendix
\thispagestyle{empty}

\onecolumn

\appendix
\section{Detailed Literature Survey}
\label{app:broad_rel_work}

\subsection{Reasoning with Uncertain Auxiliary Evidence}
\label{subsec:uncertain_evidence}
Prior work on uncertain auxiliary evidence spans formal probabilistic reasoning, belief revision under external information, and uncertainty-aware agentic systems.~\citep{nafar2025reasoning} introduce BLInD, a benchmark in which LLMs answer queries from textual premises containing explicitly quantified probabilities. Direct and chain-of-thought prompting deteriorate as the underlying Bayesian networks become more complex, whereas translating the premises into executable symbolic representations substantially improves accuracy. BLInD therefore evaluates formal probability computation and the value of symbolic execution, rather than whether uncertainty associated with naturally produced evidence should be communicated to a downstream model.

From a complementary behavioral perspective,~\citep{pmlr-v318-ghanbari26a} stratify multiple-choice predictions by the LLM's initial confidence and inject supportive, opposing, or irrelevant knowledge-graph statements. Low- and moderate-confidence predictions are more responsive to external information, particularly when it supports alternatives that the model already considers plausible. In contrast, high-confidence predictions are comparatively resistant to revision. Their analysis uses the downstream model's confidence to explain knowledge acceptance rather than studying the uncertainty reported by the evidence source.

Uncertainty has also been incorporated into multimodal agentic pipelines. SRICE~\citep{zhi2025seeing} is a training-free VQA framework that applies conformal procedures to segmentation and object-detection outputs and uses output uncertainty to select among tool-conditioned pathways. Its ablations additionally attach raw confidence scores to textualized tool evidence; this produces only marginal gains over providing tool outputs alone and underperforms the conformally calibrated pipeline. VideoAgent2~\citep{zhi2025videoagent} more directly communicates confidence by attaching uncertainty scores to observations returned by captioning, detection, and tracking tools. The LLM is instructed to use these scores when revising its retrieval plan and synthesizing an answer, and removing them reduces average accuracy across the evaluated long-video QA benchmarks. However, because confidence influences subsequent tool selection and retrieval, it can also change which evidence the model eventually observes. Consequently, the contribution of uncertainty communication cannot be separated from changes to the evidence-acquisition trajectory.

Related research examines whether external contexts should be trusted and whether all retrieved evidence is equally useful.~\citep{huang2025trust} formulate \emph{situated faithfulness} to study how LLMs arbitrate between internal knowledge and external contexts that may be correct or misleading. They show that models frequently over-rely on external information and propose self-guided and rule-based confidence-reasoning procedures for estimating contextual reliability before answer selection. These methods infer whether the external context as a whole should be trusted, rather than communicating the graded uncertainty supplied by its source. 

PANDORA~\citep{wu2024pandora} instead varies the composition of retrieved evidence across seven forms of RAG noise. Its results demonstrate that evidence items differ markedly in utility: counterfactual and other harmful contexts reduce accuracy, while some semantically irrelevant or malformed contexts can unexpectedly improve accuracy and confidence. This heterogeneity motivates examining evidence subsets rather than treating auxiliary information as uniformly useful.

\textbf{Key Differences.} In summary, these studies formalize probabilistic premises, explain knowledge acceptance using model confidence, use uncertainty to calibrate or route agentic pipelines, estimate contextual trustworthiness, or manipulate the quality of retrieved evidence. Unlike these works, our paired design holds task-native auxiliary evidence fixed and changes only whether its associated uncertainty is communicated to the foundation model; we then evaluate its effects across evidence sources, domains, models, and evidence subsets.

\subsection{Reasoning through Attributes and Intermediate Representations}
\label{subsec:intermediate_evidence}
Concept bottleneck models provide an interpretable alternative to end-to-end prediction by requiring decisions to pass through human-understandable intermediate concepts~\citep{pmlr-v119-koh20a}. This structure permits concept-level inspection and intervention, but its performance depends critically on whether the selected concepts are informative and whether they can be reliably inferred from the input. Moving beyond deterministic concept predictions, \citet{pmlr-v202-kim23g} introduce Probabilistic Concept Bottleneck Models, which represent each predicted concept through a probabilistic embedding and propagate concept uncertainty to the final class prediction. On CUB and AwA2, concept uncertainty increases when relevant visual information is obscured, correlates with lower concept and class accuracy, and supports more effective uncertainty-guided interventions. ProbCBM therefore incorporates uncertainty directly into a trained, task-specific bottleneck architecture.

The reliability of the concept layer also depends on its supervision. \citet{park2026analysis} shows that annotation noise degrades task accuracy, concept alignment, and the effectiveness of concept-level interventions. Importantly, much of this degradation is concentrated in a relatively small subset of noise-susceptible concepts rather than being distributed uniformly across the bottleneck. They use sharpness-aware minimization to improve robustness during training and predictive entropy to identify concepts to prioritize for correction at inference time. Their results demonstrate that uncertainty can expose particularly damaging intermediate representations, but the uncertainty is inferred within a trained concept model and used to repair its bottleneck under noisy supervision.

Concept-based prediction has also been extended to pretrained vision-language models. \citet{menon2023visual} introduces classification by description, in which an LLM generates class-specific visual descriptors and their image--text similarities are aggregated to classify an image. The resulting linguistic attributes provide an interpretable and editable intermediate representation while improving zero-shot recognition across several visual domains. However, the descriptor similarities function internally as components of the classifier rather than as uncertainty information explicitly presented to a downstream foundation model. Taking a Bayesian approach to concept construction, \citet{CBM2026} introduces BC-LLM, which iteratively uses an LLM to propose and annotate concepts within a Bayesian variable-selection procedure. Observed data can consequently correct unreliable LLM proposals, while posterior inference quantifies uncertainty over the resulting concept sets.

\textbf{Key Differences.} Collectively, these approaches use uncertainty to represent, train, select, evaluate, or repair an intermediate concept layer. In contrast, we begin with task-native attributes produced by external sources, hold both their content and the foundation model fixed, and vary whether their associated uncertainty is explicitly communicated. We additionally vary the supplied attribute subsets and evaluate the resulting final predictions across evidence sources, domains, and foundation models.

\subsection{Upstream and Downstream Uncertainty in Foundation Models}
\label{subsec:upstream_downstream_uncertainty}
Foundation-model uncertainty has primarily been studied as a property of the model's knowledge or of its generated responses. \citet{kadavath2022language} show that LLMs can evaluate whether their answers are correct and whether they possess the knowledge required to answer a question. Semantic entropy instead groups generations by meaning, preventing lexical variation among semantically equivalent answers from being interpreted as uncertainty~\citep{kuhn2023semantic}. In black-box settings, \citet{xiong2024can} find that directly verbalized confidence is frequently overconfident, while consistency across repeated responses provides a more informative signal of failure. Rather than evaluating uncertainty after generation, \citet{chen2026query} estimates whether a query is answerable beforehand, enabling uncertainty-aware retrieval and model cascading. These approaches characterize uncertainty originating within the downstream model.

Subsequent work further distinguishes uncertainty about knowledge from legitimate variability in model responses.~\citep{yadkori2024believe} iteratively prompt an LLM with its previous responses and measure their dependence through mutual information. This separates epistemic uncertainty from high response entropy due to multiple valid answers, thereby improving hallucination detection and calibrated abstention. ~\citep{kumaran2026causal} provide causal evidence that internal confidence influences whether an LLM answers or abstains. Through activation steering and explicit manipulation of decision thresholds, they show that log-probability-based and verbal confidence both affect abstention. However, each provides only a partial read-out of a richer internal confidence representation. Notably, verbalized confidence can influence model behavior even when it is comparatively weak at predicting correctness.

External context can itself alter the reliability of downstream uncertainty estimates. \citet{soudani-etal-2025-uncertainty} show that conventional estimators in retrieval-augmented generation often report lower uncertainty merely because contextual evidence is present, including when that evidence is irrelevant or contradicts the generated response. They formulate axioms describing how response uncertainty should behave under supportive, contradictory, and irrelevant contexts and introduce evidence-aware calibration that strengthens its association with correctness. Their uncertainty nevertheless remains a property of the downstream response after contextual augmentation. Multimodal models introduce an additional source of variation: \citet{xuan2025seeing} find that verbalized confidence in VLMs is frequently miscalibrated and sensitive to input modality, with semantically aligned visual inputs generally producing poorer calibration than corresponding textual inputs. Their confidence-aware prompting improves the calibration of the VLM's self-reported confidence but does not communicate uncertainty originating from an upstream evidence source.

A smaller body of work jointly considers uncertainty from multiple components of an augmented system. Lymperopoulos and Sarathy~\citep{tools-uncertainty,lymperopoulos2025toolsloopquantifyinguncertainty} formulate system-level uncertainty for tool-augmented question answering by combining uncertainty in the external tool with predictive or semantic uncertainty in the final LLM response. Under their strong-tool approximation, the entropy of an external classifier or retriever is added to the entropy of the final answer. The resulting measure is generally more predictive of correctness on synthetic classifier-based QA tasks, although improvements are smaller and model-dependent in their RAG proof of concept. Their objective is to assess the reliability of the completed composite system rather than expose tool uncertainty to the LLM that produces the answer. Within tool-using agents, ProbeCal learns embedding-based estimates of whether alternative prompts and execution traces will succeed, and uses these calibrated scores to select trajectories~\citep {liu-etal-2024-uncertainty}. It therefore calibrates confidence in an agent's tool-use trajectory rather than representing uncertainty associated with the resulting external evidence.
Collectively, this literature estimates downstream self-knowledge, semantic variability, query answerability, abstention confidence, context-sensitive response uncertainty, multimodal calibration, composite-system reliability, or tool-use success. 

\textbf{Key Differences.} Unlike these approaches, our intervention communicates uncertainty emitted by the original auxiliary-evidence source to an otherwise unchanged foundation model while holding the evidence itself fixed. We then evaluate its effect on final predictions, whether upstream uncertainty predicts downstream correctness, and whether upstream uncertainty is reflected in the foundation model's own predictive uncertainty across evidence sources, domains, and models.
\section{Additional Implementation Details}
\label{app:additional_implementation}

\subsection{Evaluation Protocol}
\label{app:evaluation_protocol}

\paragraph{Repeated runs and reproducibility.} All experiments are repeated three times using seeds 42, 96, and 1001. For \cub{}, the seed determines the sampled evaluation set, with 40 images drawn per class for each 7-, 8-, and 9-class configuration; paired conditions use the same sampled images within each run, and results are reported as mean $\pm$ standard deviation across seeds. In contrast, \sun{} and \dermseven{} use fixed evaluation sets of 80 and 395 images, respectively. Because inference is deterministic and the input samples remain unchanged, repeated runs can produce identical predictions, resulting in zero variance. In such cases, we report the common performance value without appending $\pm 0.0000$ for clarity. All experiments were conducted on Google Colab using NVIDIA A100-SXM4-80GB GPUs.

\paragraph{Classification performance.} Our primary evaluation metric is macro-F1, which weights each target class equally. For every paired configuration, the effect of communicating uncertainty is reported as
\[
    \Delta =
    100\left(
    \mathrm{F1}_{\auxunc{}} -
    \mathrm{F1}_{\aux{}}
    \right),
\]
in percentage points. For \dermseven{}, all nonempty subsets of the seven findings are evaluated in the text-only experiments, and results are additionally grouped by subset cardinality.

\paragraph{Decision changes and correctness.} To complement aggregate performance, we examine paired sample-level transitions between \aux{} and \auxunc{}. We distinguish cases in which communicating uncertainty changes an incorrect prediction to a correct one from those in which it changes a correct prediction to an incorrect one. We additionally evaluate whether upstream uncertainty is informative of downstream prediction correctness using AUROC. Because the available uncertainty has different semantics across \cub{}, \sun{}, and \dermseven{}, the corresponding sample-level score construction is defined separately for each evidence source in Appendix~\ref{app:uncertainty_features}.

\subsection{Additional Dataset Details}

\subsubsection{CUB Candidate-Class Construction}
\label{app:cub_classes}
\begin{table}[t]
\centering
\small
\caption{Nested candidate-species configurations used for CUB. Each larger configuration retains all species from the preceding setting.}
\label{tab:cub_candidate_classes}
\begin{tabular}{cl}
\toprule
\textbf{Configuration} & \textbf{Candidate species added} \\
\midrule
3-class &
Chuck-will's-widow (22), Green Jay (75), White Pelican (101) \\
5-class &
+ Whip-poor-will (105), Hooded Warbler (167) \\
7-class &
+ Ivory Gull (63), Painted Bunting (16) \\
8-class &
+ Nighthawk (92) \\
9-class &
+ Herring Gull (62) \\
\bottomrule
\end{tabular}
\end{table}
Rather than selecting candidate species independently for each class count, we construct a nested sequence of increasingly ambiguous classification problems. We represent each of the 200 CUB species by its continuous 312-dimensional class-level attribute profile and use cosine similarity in this attribute space to guide candidate selection.

We begin with a deliberately well-separated three-class configuration. Among the 200 species, we select \textit{Chuck-will's-widow} (class 22), \textit{Green Jay} (75), and \textit{White Pelican} (101) by maximizing the minimum pairwise cosine distance between their class-level attribute profiles. We then progressively enlarge this fixed set by introducing species whose attribute profiles are more similar to existing candidates. The five-class configuration adds \textit{Whip-poor-will} (105) and \textit{Hooded Warbler} (167). The seven-class configuration further adds \textit{Ivory Gull} (63) and \textit{Painted Bunting} (16). These additions introduce interpretable competing groups: \textit{Whip-poor-will} is highly similar to \textit{Chuck-will's-widow}, \textit{Hooded Warbler} and \textit{Painted Bunting} provide closer attribute-space alternatives to \textit{Green Jay}, and \textit{Ivory Gull} is close to \textit{White Pelican}.

Finally, we extend the seven-class set with \textit{Nighthawk} (92) and \textit{Herring Gull} (62). \textit{Nighthawk} creates a closely related attribute-space group with \textit{Whip-poor-will} and \textit{Chuck-will's-widow}. \textit{Herring Gull} similarly introduces ambiguity around the light-coloured water-bird candidates, with similarities to \textit{Ivory Gull} and \textit{White Pelican}. The resulting eight- and nine-class configurations, therefore, increase candidate ambiguity while preserving all classes from the smaller configurations. The list of classes used in different configurations is summarized in \tablename~\ref{tab:cub_candidate_classes}.

For each image $i$, we retain only visual attributes annotated as present and exclude those marked as not visible. These retained observations constitute the auxiliary evidence $\mathbf{a}_i$. Each evidence $a_i^j$ is accompanied by the annotator-reported ordinal certainty signal $u_i^j\in\{\textit{guessing},\textit{probably},\textit{definitely}\}$. For each retained attribute, we also provide its continuous class-level score for each candidate species.

For a fixed image and candidate configuration, the auxiliary-only and uncertainty-aware conditions share the same retained attributes and class-level attribute profiles. The only difference is that the uncertainty-aware condition additionally communicates the corresponding human certainty label $u_i^j$ for each retained evidence $a_i^j$. This construction, therefore, isolates the effect of communicating the available human certainty while holding the underlying auxiliary evidence fixed.

\subsubsection{SUN Scene and Attribute Construction}
\label{app:sun_construction}

\paragraph{SUN attribute representation.}
The SUN Attribute Database provides 102 scene attributes together with Amazon Mechanical Turk annotation vote fractions. We retain these vote fractions as continuous values rather than thresholding them into binary labels. In the original annotations, the per-image vote fractions take values in $\{0,\frac{1}{3},\frac{2}{3},1\}$, corresponding to agreement among three annotators. For each candidate scene category, we construct a class-level attribute profile by averaging these continuous vote fractions over images belonging to that class.

\paragraph{Attribute selection.}
Starting from the 102 SUN attributes, the selection procedure first removes attributes that are rare within the selected scene classes, defined as being present in fewer than 5\% of images. For the remaining attributes, discriminativeness is estimated using two complementary statistics: mutual information between the attribute values and the scene category, and the maximum absolute coefficient obtained from an $\ell_1$-regularised multinomial logistic-regression classifier. Each attribute is ranked independently according to these two criteria, and the two ranks are averaged to obtain a combined ranking. The highest-ranked candidates are then retained subject to an attribute budget of approximately 20.

The implementation additionally permits selected correlated attribute groups to be retained deliberately rather than automatically removing redundant features. In particular, it defines the pairs \{\textit{open area}, \textit{enclosed area}\}, \{\textit{diving}, \textit{sunbathing}\}, and \{\textit{ocean}, \textit{sailing/boating}\}; a group is protected only when all of its members are non-rare for the selected scene classes. This produces the 20-attribute pool used for subsequent BLIP scoring.

\paragraph{Image-specific BLIP scores.}
We use the pretrained BLIP image--text-matching model \texttt{Salesforce/blip-itm-base-coco}. For each image and each of the 20 selected attributes, we construct a textual description of the form ``a photo with [attribute]''. Attribute names are lightly normalised by removing parenthetical qualifiers and replacing ``/'' with ``or'' before forming the caption. All 20 captions are evaluated against the same image using BLIP's ITM head, and we record
\[
    s_{i,a}
    =
    \operatorname{softmax}
    \bigl(\mathrm{ITM}(x_i,a)\bigr)_{\mathrm{match}},
\]
where $s_{i,a}\in[0,1]$ is the score assigned to the ITM match class. The scores are stored exactly as produced by BLIP, without thresholding or subsequent calibration.

\paragraph{Top-$K$ auxiliary evidence.}
For every image, the 20 attributes are sorted in descending order of $s_{i,a}$. We retain the first $K\in\{3,5,8,10\}$ attributes as the image-specific auxiliary evidence $\mathbf{a}_i$. For each retained evidence $a_i^j$, its corresponding BLIP match score is used as the uncertainty signal $u_i^j$. Consequently, increasing $K$ produces nested evidence sets for a fixed image: the top-3 attributes are contained in the top-5 set, which is contained in the top-8 and top-10 sets. Ties are resolved deterministically by attribute name. The same selected attributes and ordering are used in both experimental conditions. In \aux{}, the downstream model observes each attribute together with its candidate-class profile; in \auxunc{}, the identical evidence is shown together with its BLIP match score. Hence, the paired comparison does not alter which attributes are supplied, and isolates the effect of making the underlying score magnitude explicit.

\paragraph{Scene categories.}
We evaluate four SUN scene categories: \textit{park}, \textit{office cubicles}, \textit{supermarket}, and \textit{coffee shop}. These categories were specified explicitly as the candidate set prior to attribute selection; the automated preprocessing pipeline does not select the scene categories. Given this fixed class set, the subsequent procedure selects the visual and contextual attributes used for auxiliary evidence.


\subsubsection{Derm7pt Auxiliary-Evidence Construction}
\label{app:derm7pt_setup}

\paragraph{Seven-point checklist findings.} For \dermseven{}, the auxiliary evidence is derived from the seven dermatological findings in the seven-point checklist: \textit{pigment network}, \textit{blue-whitish veil}, \textit{vascular structures}, \textit{pigmentation}, \textit{streaks}, \textit{dots and globules}, and \textit{regression structures}. The original categorical annotation for each finding is mapped to a binary state indicating whether the finding is considered \textit{not suspicious} or \textit{suspicious}. The mappings used throughout the experiments are listed in Table~\ref{tab:derm7pt_binary_mapping}.

\begin{table}[t]
\centering
\small
\caption{Binarisation of the original Derm7pt checklist annotations used to
train the seven auxiliary finding predictors.}
\label{tab:derm7pt_binary_mapping}
\begin{tabular}{p{0.22\linewidth}p{0.34\linewidth}p{0.34\linewidth}}
\toprule
\textbf{Finding} &
\textbf{Not suspicious} &
\textbf{Suspicious} \\
\midrule
Pigment network &
absent; typical &
atypical \\

Blue-whitish veil &
absent &
present \\

Vascular structures &
absent; arborizing; comma; hairpin; within regression; wreath &
dotted; linear irregular \\

Pigmentation &
absent; diffuse regular; localized regular &
diffuse irregular; localized irregular \\

Streaks &
absent; regular &
irregular \\

Dots and globules &
absent; regular &
irregular \\

Regression structures &
absent &
blue areas; white areas; combinations \\
\bottomrule
\end{tabular}
\end{table}

\paragraph{Upstream finding predictors.}
We train one independent \resnet{} classifier for each of the seven binary checklist findings using the Derm7pt training split. All classifiers operate on the clinical-image modality and are initialized from ImageNet-pretrained weights. The original ResNet classification head is replaced with a two-class linear classifier corresponding to \textit{not suspicious} and \textit{suspicious}. Images are resized to $224\times224$ pixels and normalized using ImageNet statistics. During training, we additionally apply random cropping, horizontal and vertical flipping, rotation, color jitter, and random erasing. To mitigate the substantial class imbalance across several findings, training uses inverse-frequency-weighted sampling.

Each classifier is trained for 20 epochs using Adam with a learning rate of $10^{-4}$, weight decay of $10^{-4}$, a batch size of 32, cross-entropy loss, and a cosine learning-rate schedule. For each finding, the checkpoint attaining the highest validation macro-F1 is retained. Held-out performance of the resulting seven classifiers on the common 395-image test split is reported in Table~\ref{tab:derm7pt_aux_resnet}. The models differ substantially in their ability to recover the individual findings, with test macro-F1 ranging from $0.5612$ to $0.7078$. This variation is useful for our downstream study because the auxiliary evidence is therefore neither uniformly reliable nor artificially noise-free.

For an input image $\mathbf{x}_i$, predictor $g_j$ produces a probability vector
\[
    \mathbf{p}_{ij}
    =
    \operatorname{softmax}\!\left(g_j(\mathbf{x}_i)\right).
\]
The predicted binary finding is
\[
    a_i^j
    =
    \arg\max_c p_{ijc},
\]
and its corresponding maximum-softmax confidence is
\[
    u_i^j
    =
    \max_c p_{ijc}.
\]
We retain this confidence exactly as produced by the upstream classifier and do not apply additional calibration. Consequently, \aux{} communicates the predicted binary findings, while \auxunc{} communicates the identical findings together with their associated confidence scores.

\paragraph{Downstream target and reference evidence.}
The downstream task collapses the Derm7pt diagnosis into a binary target, with melanoma and basal cell carcinoma treated as malignant and all remaining diagnoses treated as non-malignant. We evaluate the fixed test split of 395 clinical images, containing 117 malignant and 278 non-malignant cases. The same images are used to obtain the predicted findings and confidence scores from the seven upstream classifiers.

In addition to these predicted-evidence conditions, we construct two reference conditions directly from the ground-truth checklist annotations. \goldbinary{} applies the same binary mappings in Table~\ref{tab:derm7pt_binary_mapping}, whereas \goldoriginal{} retains the original categorical annotations. These conditions remove errors introduced by the seven upstream predictors and therefore help distinguish limitations associated with upstream evidence quality from those of the downstream foundation model.
\begin{table}[t]
\centering
\small
\caption{Held-out test performance of the seven Derm7pt auxiliary ResNet classifiers. Each model predicts whether the corresponding dermoscopic finding is \texttt{not\_susp} or \texttt{suspicious}. All results are computed on the same 395 clinic images from the fixed test split. Macro-F1 is the primary metric; class-wise F1 and accuracy are also reported.}
\label{tab:derm7pt_aux_resnet}
\begin{tabular}{lcccc}
\toprule
\textbf{Auxiliary Finding}
& \textbf{Macro-F1}
& \textbf{F1 Not-Susp.}
& \textbf{F1 Suspicious}
& \textbf{Accuracy} \\
\midrule
Blue-whitish veil      & \textbf{0.7078} & 0.8528 & 0.5628 & 0.7797 \\
Regression structures & 0.6720 & 0.8148 & 0.5291 & 0.7342 \\
Dots and globules      & 0.6063 & 0.5674 & \textbf{0.6452} & 0.6101 \\
Pigmentation           & 0.5949 & 0.7231 & 0.4667 & 0.6354 \\
Vascular structures    & 0.5852 & \textbf{0.9370} & 0.2333 & \textbf{0.8835} \\
Pigment network        & 0.5786 & 0.8218 & 0.3353 & 0.7190 \\
Streaks                & 0.5612 & 0.6734 & 0.4490 & 0.5899 \\
\bottomrule
\end{tabular}
\end{table}

\subsection{Downstream Generative Foundation Models and Inference}
\label{app:downstream_models}

\paragraph{Large language models.}
We evaluate \qwenllm{}, \llamallm{}, and \phillm{} using the checkpoints \texttt{Qwen/Qwen3-32B}, \texttt{meta-llama/Llama-3.1-8B-Instruct}, and \texttt{microsoft/phi-4}, respectively. All models are evaluated without fine-tuning using their native chat templates and zero-shot task prompts. For \qwenllm{}, we disable its optional thinking mode using \texttt{enable\_thinking=False}; none of the models is explicitly prompted to produce chain-of-thought reasoning.

The models are loaded in \texttt{bfloat16} precision on CUDA with automatic device placement and evaluated in inference mode. Decoding is deterministic (\texttt{do\_sample=False}), without temperature, top-$p$, or top-$k$ sampling. Generation is limited to 32 tokens for \dermseven{} and 64 tokens for \cub{} and \sun{}. Each prompt requests a single class identifier, and raw model responses and sample-level predictions are retained for subsequent analysis.

\paragraph{Vision--language models.}
For the multimodal evaluation, we use \qwenvl{} and \internvl{}. Both models receive the original image together with the textual task context. We evaluate the \imageonly{}, \imageaux{}, and \imageauxunc{} conditions; for \dermseven{}, we additionally evaluate the \goldbinary{} and \goldoriginal{} reference conditions.

Both VLMs are evaluated without fine-tuning using model-specific image preprocessing and deterministic decoding. For \qwenvl{}, the processor uses a pixel-count budget ranging from $256\times28^2$ to $1024\times28^2$, with generation limited to 64 tokens. \internvl{} uses $448\times448$ dynamic image tiles with at most 12 tiles. Since the classification tasks require a single candidate identifier, the final \internvl{} evaluations use \texttt{max\_new\_tokens=1}. Image preprocessing and all inference settings are held fixed across paired evidence conditions for each model.

\subsection{Correctness Probe}
\label{app:correctness_probe}

To evaluate whether upstream uncertainty is informative about downstream decision correctness, we train a lightweight logistic regression probe on uncertainty-derived features and predict whether the downstream decision is correct. We use the same probing protocol across all benchmarks and model configurations.

Features are standardized using \texttt{StandardScaler} and passed to an $\ell_2$-regularized logistic-regression model with $C=1$, the \texttt{liblinear} solver, balanced class weights, and a maximum of 2000 iterations. Evaluation uses five-fold stratified cross-validation with shuffling and random seed 42. Standardization is fitted independently within each training fold through a pipeline to avoid leakage from the held-out fold.

For each sample, the probe outputs a probability that the downstream decision is correct. We concatenate the out-of-fold probabilities across the five folds and report the pooled AUROC as the primary metric. Separate probes are trained for each benchmark, model, inference condition, and uncertainty representation; samples are never pooled across these settings.

\subsection{Uncertainty Features}
\label{app:uncertainty_features}

\begin{table}[t]
\centering
\footnotesize
\caption{Upstream features used by the downstream correctness probes. Features are constructed solely from the auxiliary evidence and its associated certainty or confidence signals, independently of the downstream model's predictions.}
\label{tab:correctness_features}
\begin{tabular}{@{}p{0.33\linewidth}p{0.63\linewidth}@{}}
\toprule
\textbf{Feature / representation} & \textbf{Definition / interpretation} \\
\midrule

\multicolumn{2}{@{}l}{\textit{CUB: Human-reported certainty}} \\

\texttt{frac\_guessing}
& Fraction of retained attributes annotated as \textit{guessing}. \\

\texttt{frac\_probably}
& Fraction of retained attributes annotated as \textit{probably}. \\

\texttt{frac\_definitely}
& Fraction of retained attributes annotated as \textit{definitely}. \\

\texttt{mean\_certainty}
& Mean ordinal certainty using the original certainty IDs
$2$, $3$, and $4$. \\

\texttt{certainty\_entropy}
& Normalized entropy of the three certainty-level fractions:
$-\sum_{c=2}^{4} f_c\log(f_c)/\log 3$. \\

\texttt{certainty\_concentration}
& Largest fraction among the three certainty levels,
$\max_c f_c$. \\

\texttt{num\_attributes}
& Number of retained attributes, capturing the amount of available
auxiliary evidence. \\

\midrule
\multicolumn{2}{@{}l}{\textit{SUN: BLIP image--text matching scores}} \\

\texttt{mean\_blip}
& Mean BLIP matching score across the selected top-$K$ attributes. \\

\texttt{min\_blip}
& Lowest BLIP matching score among the selected attributes. \\

\texttt{top1\_topk\_gap}
& Difference between the highest and lowest selected matching scores. \\

\texttt{std\_blip}
& Standard deviation of the selected BLIP matching scores. \\

\midrule
\multicolumn{2}{@{}l}{\textit{Derm7pt: Specialised classifier predictive distributions}} \\

\textbf{Confidence}
& Seven-dimensional vector containing the maximum-softmax confidence
$u_i^j=\max_c p_{ijc}$ for each checklist finding $j$. \\

\textbf{Entropy}
& Seven-dimensional vector containing the binary predictive entropy
$h_i^j$ of each finding classifier. \\

\textbf{Full}
& Twenty-one-dimensional representation containing
$[p_{i,j,0},p_{i,j,1},h_i^j]$ for each of the seven findings, where
$p_{j,0}$ and $p_{j,1}$ denote the probabilities of
\textit{not suspicious} and \textit{suspicious}, respectively. \\

\bottomrule
\end{tabular}
\end{table}

To assess whether upstream uncertainty provides information about downstream decision correctness, we construct dataset-specific features directly from the auxiliary evidence and its associated uncertainty signals. For CUB, these features capture the distribution and concentration of human-reported certainty across retained attributes, along with evidence quantity. For SUN, they summarize the strength and dispersion of BLIP image--text matching scores, while Derm7pt uses confidence from the specialized classifiers associated with the selected findings. All features are independent of the downstream LLM's predictions, token probabilities, and output confidence. \tablename~\ref{tab:correctness_features} provides their complete definitions.

\subsection{Obtaining Predictive Uncertainty of the Downstream Generative Foundation Model}
\label{app:downstream_uncertainty}

For \dermseven{}, we measure the downstream model's predictive uncertainty from its probability distribution over the two valid classes~\citep{kuhn2023semantic}, $\mathcal{Y}=\{\textit{malignant},\textit{non-malignant}\}$. Let $q_i(y)$ denote the probability assigned to class $y$ after normalizing the probability mass over these two alternatives. For the LLM, we obtain $q_i(y)$ from the token probabilities assigned to the two valid class labels. For the VLM, we use the logits for the first generated token corresponding to the two candidate classes and normalize them over the same set of alternatives. We then compute predictive entropy as
\[
H_i=-\sum_{y\in\mathcal{Y}}q_i(y)\log q_i(y),
\]
where a larger $H_i$ indicates greater downstream uncertainty, the same procedure is applied under both \aux{} and \auxunc{}; only the context supplied to the downstream model changes.

\section{Additional Experiments}
\label{app:additional_exps}
\subsection{Experiments across Different Subset Sizes}
\label{app:subset_results}
\subsubsection{SUN LLM Results}
The extended SUN results in Table~\ref{tab:sun_topk} show that the benefit of communicating uncertainty is generally preserved across different evidence budgets. The effect is strongest for \qwenllm{}, particularly as more attributes are included, while \llamallm{} and \phillm{} show smaller but mostly positive gains. The near-zero change for \llamallm{} at Top-5 is the only setting where uncertainty does not help. Overall, the results indicate that the value of uncertainty depends on both the downstream model and the amount of auxiliary evidence available.
\begin{table}[t]
\centering
\small
\caption{SUN macro-F1 across top-$K$ auxiliary-attribute settings. $\Delta$ denotes the \auxunc{} gain over \aux{} in percentage points.}
\label{tab:sun_topk}
\begin{tabular}{lcccr}
\toprule
\textbf{Model} &
\textbf{Top-$K$} &
\textbf{Aux} &
\textbf{Aux + Uncertainty} &
$\boldsymbol{\Delta}$ \textbf{(pp)} \\
\midrule

\multirow{4}{*}{Qwen3-32B}
& 3  & 0.6320 & 0.7028 & +7.09 \\
& 5  & 0.7181 & 0.7398 & +2.17 \\
& 8  & 0.6728 & 0.8690 & +19.62 \\
& 10 & 0.6066 & 0.8684 & \textbf{+26.18} \\

\midrule

\multirow{4}{*}{Llama-3.1-8B}
& 3  & 0.8881 & 0.9244 & +3.63 \\
& 5  & 0.8892 & 0.8874 & -0.18 \\
& 8  & 0.8651 & 0.9120 & +4.69 \\
& 10 & 0.8648 & 0.9241 & +5.93 \\

\midrule

\multirow{4}{*}{Phi-4-14B}
& 3  & 0.8542 & 0.8843 & +3.02 \\
& 5  & 0.8438 & 0.8466 & +0.28 \\
& 8  & 0.8589 & 0.8849 & +2.61 \\
& 10 & 0.8175 & 0.8843 & +6.69 \\

\bottomrule
\end{tabular}
\end{table}

\subsubsection{Derm7pt LLM Results}
\begin{table*}[t]
\centering
\small
\caption{Derm7pt classification performance across evidence subset sizes. Values are the mean $\pm$ standard deviation of macro-F1 across distinct subsets; for $k=7$, the single-subset result is reported without deviation. $\Delta$ denotes Aux+Unc minus Aux in percentage points. `Gold Original' refers to the original ground-truth auxiliary findings. `Gold Binary' refers to the binarized version of the gold auxiliary findings.}
\label{tab:derm7pt_subset_results}
\begin{tabular}{llrccccc}
\toprule
Model & $k$ & \# subsets & Aux & Aux+Unc & Gold Binary & Gold Original & $\Delta$ (pp) \\
\midrule
Qwen3-32B & 1 & 7 & 0.567 $\pm$ 0.040 & 0.567 $\pm$ 0.049 & 0.655 $\pm$ 0.063 & 0.636 $\pm$ 0.106 & +0.00 \\
Qwen3-32B & 3 & 35 & 0.573 $\pm$ 0.043 & 0.595 $\pm$ 0.042 & 0.681 $\pm$ 0.062 & 0.715 $\pm$ 0.029 & +2.20 \\
Qwen3-32B & 5 & 21 & 0.581 $\pm$ 0.029 & 0.612 $\pm$ 0.012 & 0.731 $\pm$ 0.023 & 0.758 $\pm$ 0.028 & +3.14 \\
Qwen3-32B & 7 & 1 & 0.505 & 0.596 & 0.669 & 0.750 & +9.10 \\
\midrule
Llama-3.1-8B & 1 & 7 & 0.496 $\pm$ 0.086 & 0.498 $\pm$ 0.096 & 0.550 $\pm$ 0.138 & 0.542 $\pm$ 0.141 & +0.12 \\
Llama-3.1-8B & 3 & 35 & 0.557 $\pm$ 0.047 & 0.577 $\pm$ 0.050 & 0.641 $\pm$ 0.066 & 0.659 $\pm$ 0.094 & +2.05 \\
Llama-3.1-8B & 5 & 21 & 0.584 $\pm$ 0.032 & 0.619 $\pm$ 0.012 & 0.714 $\pm$ 0.024 & 0.739 $\pm$ 0.034 & +3.56 \\
Llama-3.1-8B & 7 & 1 & 0.601 & 0.630 & 0.740 & 0.805 & +2.90 \\
\midrule
Phi-4-14B & 1 & 7 & 0.567 $\pm$ 0.040 & 0.579 $\pm$ 0.040 & 0.655 $\pm$ 0.063 & 0.658 $\pm$ 0.057 & +1.19 \\
Phi-4-14B & 3 & 35 & 0.536 $\pm$ 0.048 & 0.571 $\pm$ 0.042 & 0.671 $\pm$ 0.035 & 0.705 $\pm$ 0.021 & +3.48 \\
Phi-4-14B & 5 & 21 & 0.528 $\pm$ 0.039 & 0.568 $\pm$ 0.022 & 0.638 $\pm$ 0.042 & 0.683 $\pm$ 0.024 & +4.00 \\
Phi-4-14B & 7 & 1 & 0.544 & 0.547 & 0.684 & 0.653 & +0.25 \\
\bottomrule
\end{tabular}
\end{table*}
The extended \dermseven{} results in Table~\ref{tab:derm7pt_subset_results} show that communicating uncertainty is generally more beneficial as richer evidence subsets are provided, although the magnitude depends on the model and composition. The gains are small for single findings and become larger for intermediate or full evidence sets, particularly for \qwenllm{} and \llamallm{}. The gold-reference conditions remain substantially stronger than the predicted-evidence conditions, indicating that upstream finding quality is still an important bottleneck even when its confidence is communicated.
\subsubsection{Derm7pt VLM Results}
\label{app:dermseven_vlm_results}
\begin{table*}[t]
\centering
\small
\caption{Derm7pt VLM performance across auxiliary-evidence subset sizes. All conditions include the original clinical image. \imageaux{} provides predicted checklist findings, while \imageauxunc{} provides the same findings together with their corresponding upstream classifier confidence. \goldbinary{} and \goldoriginal{} replace the predicted findings with their ground-truth binarized and original categorical annotations, respectively. Results are macro-F1 on the fixed 395-image test set. $\Delta$ denotes the paired change from \imageaux{} to \imageauxunc{} in percentage points.}
\label{tab:derm7pt_vlm}
\resizebox{\textwidth}{!}{
\begin{tabular}{lccccccc}
\toprule
\textbf{Model}
& \textbf{$k$}
& \textbf{\imageonly{}}
& \textbf{\imageaux{}}
& \textbf{\imageauxunc{}}
& \textbf{\goldbinary{}}
& \textbf{\goldoriginal{}}
& $\boldsymbol{\Delta}$ \textbf{(pp)} \\
\midrule

\multirow{4}{*}{\qwenvl{}}
& 1 & \multirow{4}{*}{0.2649} & 0.5904 & 0.5673 & 0.6655 & 0.6535 & $-2.31$ \\
& 3 &  & 0.5505 & 0.5496 & 0.6710 & 0.7017 & $-0.09$ \\
& 5 &  & 0.5422 & 0.5483 & 0.6662 & 0.7273 & $+0.61$ \\
& 7 &  & 0.5264 & 0.5442 & 0.6271 & 0.7306 & $+1.78$ \\
\midrule

\multirow{4}{*}{\internvl{}}
& 1 & \multirow{4}{*}{0.4604} & 0.4808 & 0.5301 & 0.5213 & 0.5946 & $+4.93$ \\
& 3 &  & 0.5210 & 0.5674 & 0.6315 & 0.6994 & $+4.64$ \\
& 5 &  & 0.5021 & 0.5296 & 0.6355 & 0.7014 & $+2.75$ \\
& 7 & & 0.4739 & 0.4913 & 0.6377 & 0.7220 & $+1.74$ \\
\bottomrule
\end{tabular}
}
\end{table*}
The extended \dermseven{} VLM results in Table~\ref{tab:derm7pt_vlm} show a more model-dependent effect of uncertainty communication than for the LLMs. \internvl{} benefits across all evidence sizes, whereas \qwenvl{} improves only for the larger subsets. In both models, the gold-reference conditions are substantially stronger than the predicted-evidence conditions, again indicating that upstream finding quality remains a major limitation even when confidence is communicated. The remaining gap to the gold-reference conditions also motivates future evaluation with domain-specialized medical VLMs, which may exploit dermatological evidence differently from the general-purpose models considered here.
\subsection{Analysis of Samples That Change with Additional Evidence for \sun{} Scene Classification}
\label{app:sun_attr_structure}

The preceding analysis shows that communicating BLIP scores can prevent a correct prediction from being overturned when lower-ranked attributes are added. We next ask what characterizes these cases. We focus on the transition from $K=3$ to $K=10$. For each image, we call its three highest-ranked attributes the \emph{original evidence} and attributes ranked 4--10 the \emph{added evidence}. We then measure the strength of the relationship between the two sets of attributes using the continuous SUN attribute annotation matrix.

Importantly, the correlation is not computed within a single image. Each SUN attribute defines a vector of annotation strengths across images, and we compute the Pearson correlation between these vectors. For example, suppose that for one test image the original top-3 attributes include \textit{trees}, \textit{grass}, and \textit{sky}, while the added evidence contains attributes such as \textit{road}, \textit{bench}, and \textit{building}. We ask whether, across SUN images, scenes containing \textit{trees} also tend to contain \textit{road}, whether \textit{grass} co-occurs with \textit{bench}, and so forth. With three original and seven added attributes, this gives $3\times7=21$ pairwise correlations for each test image.

We summarize these relationships in two complementary ways. The first is the mean absolute correlation,
\begin{equation}
    C_{\mathrm{abs}}
    =
    \frac{1}{21}
    \sum_{j=1}^{3}\sum_{k=4}^{10}
    \left|\rho(a^j,a^k)\right|.
\end{equation}
A value approaching $1$ would indicate that the newly added attributes are strongly associated with the original evidence, regardless of whether the association is positive or negative; a value near $0$ indicates that the two sets are largely unrelated. We additionally compute the signed mean
\begin{equation}
    C_{\mathrm{signed}}
    =
    \frac{1}{21}
    \sum_{j=1}^{3}\sum_{k=4}^{10}
    \rho(a^j,a^k).
\end{equation}
Here, a large positive value means that the added attributes generally co-occur with the original evidence, a strongly negative value indicates an opposing association, and a value near zero indicates little net alignment (or a mixture of positive and negative relationships). Finally, we compute the mean BLIP match score of the seven newly introduced attributes to measure how strongly BLIP itself associates them with the image.
\begin{wrapfigure}{r}{0.45\columnwidth}
    \centering
    \vspace{-0.8em}
    \includegraphics[width=0.75\linewidth]{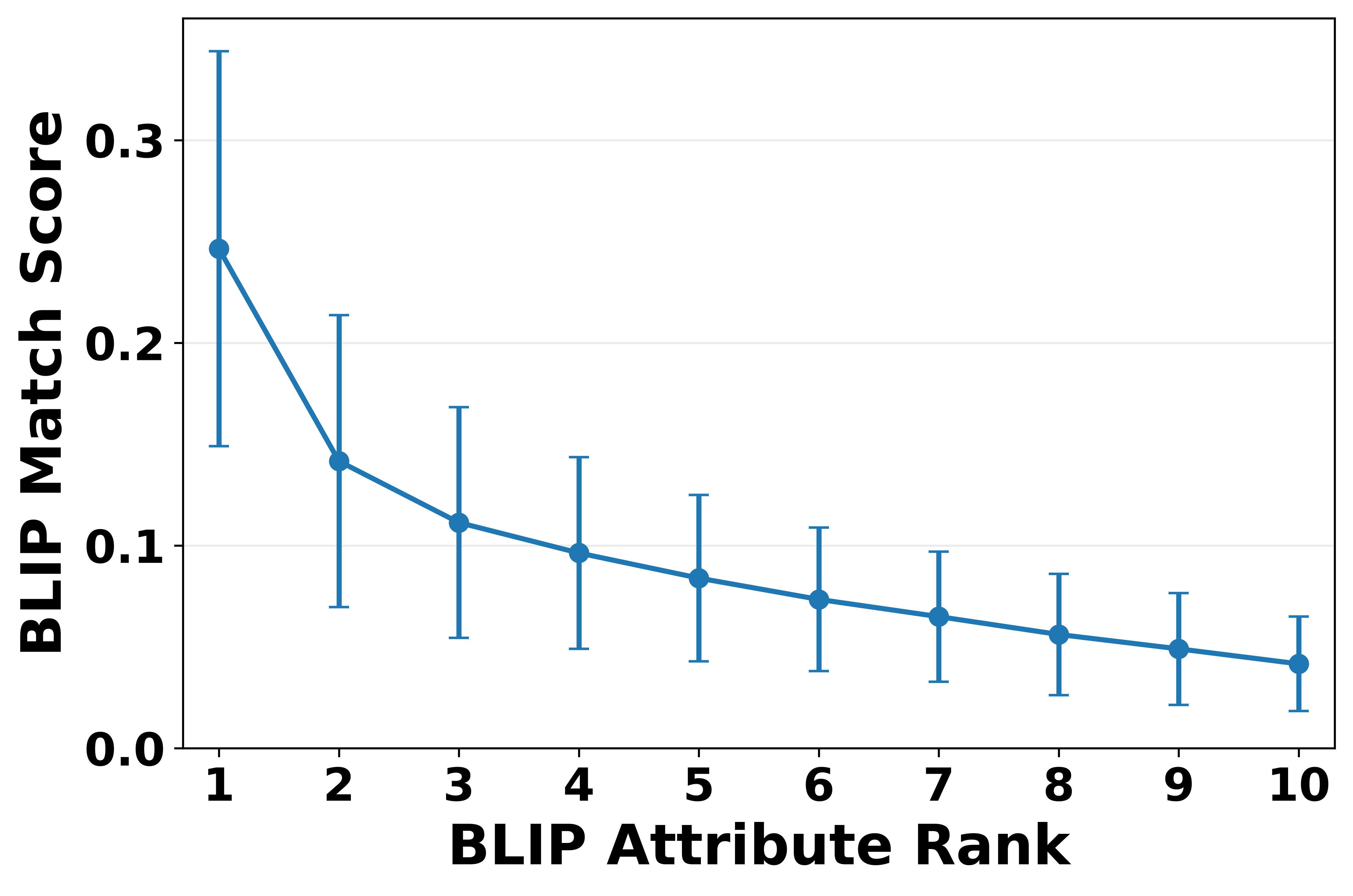}
    \caption{BLIP match score as a function of attribute rank across the 80
    selected \sun{} images. For each image, the 20 candidate attributes are
    ranked by BLIP match score; points show the mean score at each rank and
    error bars denote one standard deviation. Scores decrease with rank,
    confirming that increasing top-$K$ progressively introduces attributes
    that BLIP itself scores as less strongly matched to the image.}
    \label{fig:sun_rank_scores}
    \vspace{-0.8em}
\end{wrapfigure}

We group images by how the downstream prediction changes as $K$ increases from $K=3$ to $K=10$: correct$\rightarrow$correct, correct$\rightarrow$incorrect, incorrect$\rightarrow$correct, or incorrect$\rightarrow$incorrect. Under \aux{}, images that remain correct show the strongest association between the original and added evidence, with mean $C_{\mathrm{abs}}$ values of $0.236$, $0.219$, and $0.218$ for \qwenllm{}, \llamallm{}, and \phillm{}, respectively. Thus, predictions are more likely to remain stable when the newly introduced attributes are structurally related to the evidence already available.

We then examine the particularly informative cases that are correct$\rightarrow$incorrect under \aux{} but remain correct$\rightarrow$correct under \auxunc{}---that is, cases in which adding the seven attributes destabilizes the prediction when their scores are omitted, but not when their scores are communicated. The added attributes in these cases receive very low mean BLIP scores ($0.041$, $0.041$, and $0.051$ for \qwenllm{}, \llamallm{}, and \phillm{}) and have weak signed association with the original top-3 evidence ($C_{\mathrm{signed}}=0.042$, $0.046$, and $0.037$, respectively). Although these groups are small, they provide a consistent explanation for the robustness pattern in Table~\ref{tab:sun_robustness}: uncertainty communication is particularly useful when the newly introduced evidence is weakly scored and only loosely aligned with the stronger evidence already present.
\section{Prompt Templates}
\label{app:prompt_templates}
\begin{figure*}[t]
    \centering
    \includegraphics[width=\textwidth, keepaspectratio]{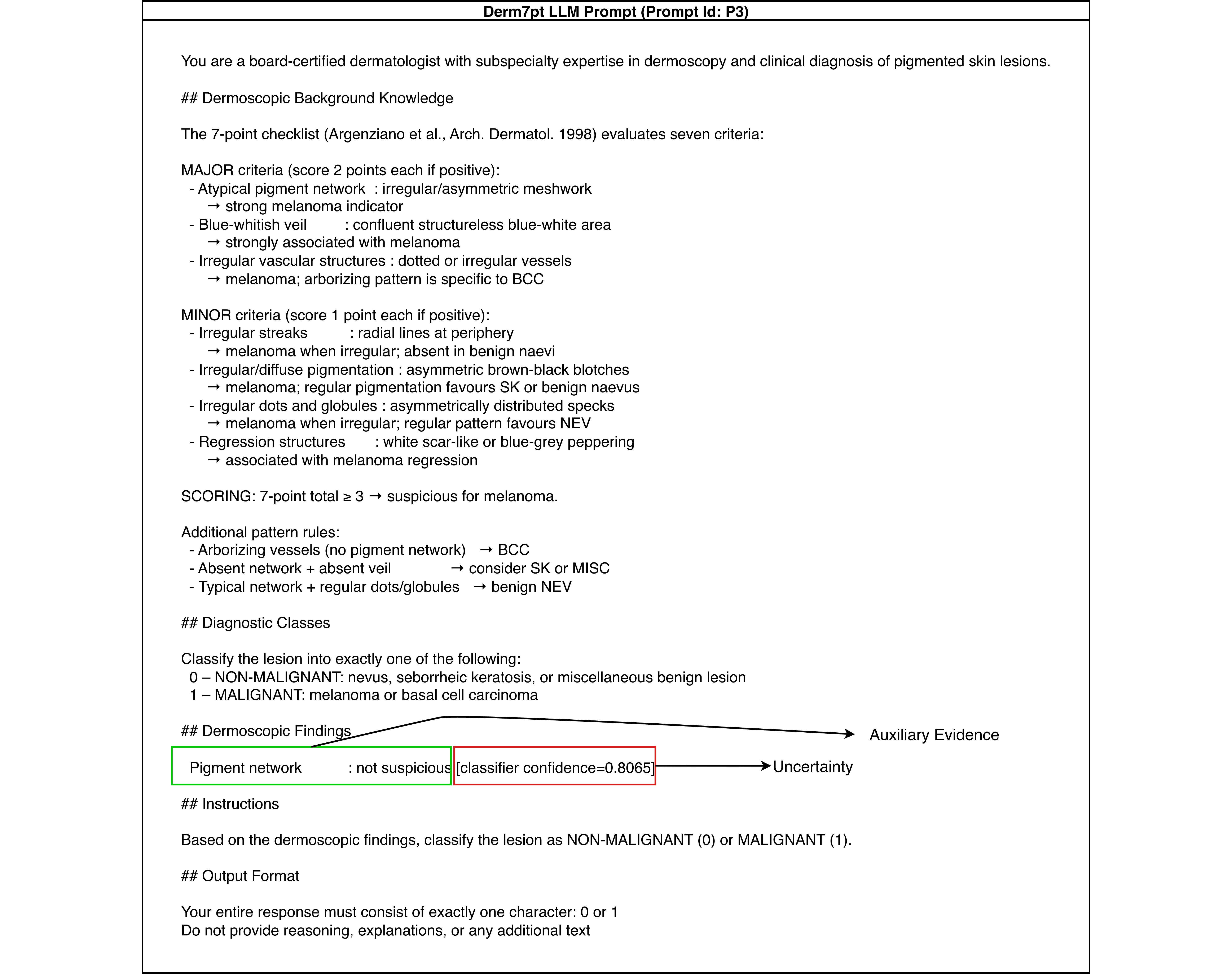}
    \caption{\dermseven{} LLM Prompt. This is an illustrative example with only one auxiliary evidence and its associated classifier confidence. The prompt contains additional background information obtained from~\citep{argenziano1998epiluminescence}.}
    \label{fig:derm7pt_llm_final_prompt}
\end{figure*}

\begin{figure*}[t]
    \centering
    \includegraphics[width=\textwidth, keepaspectratio]{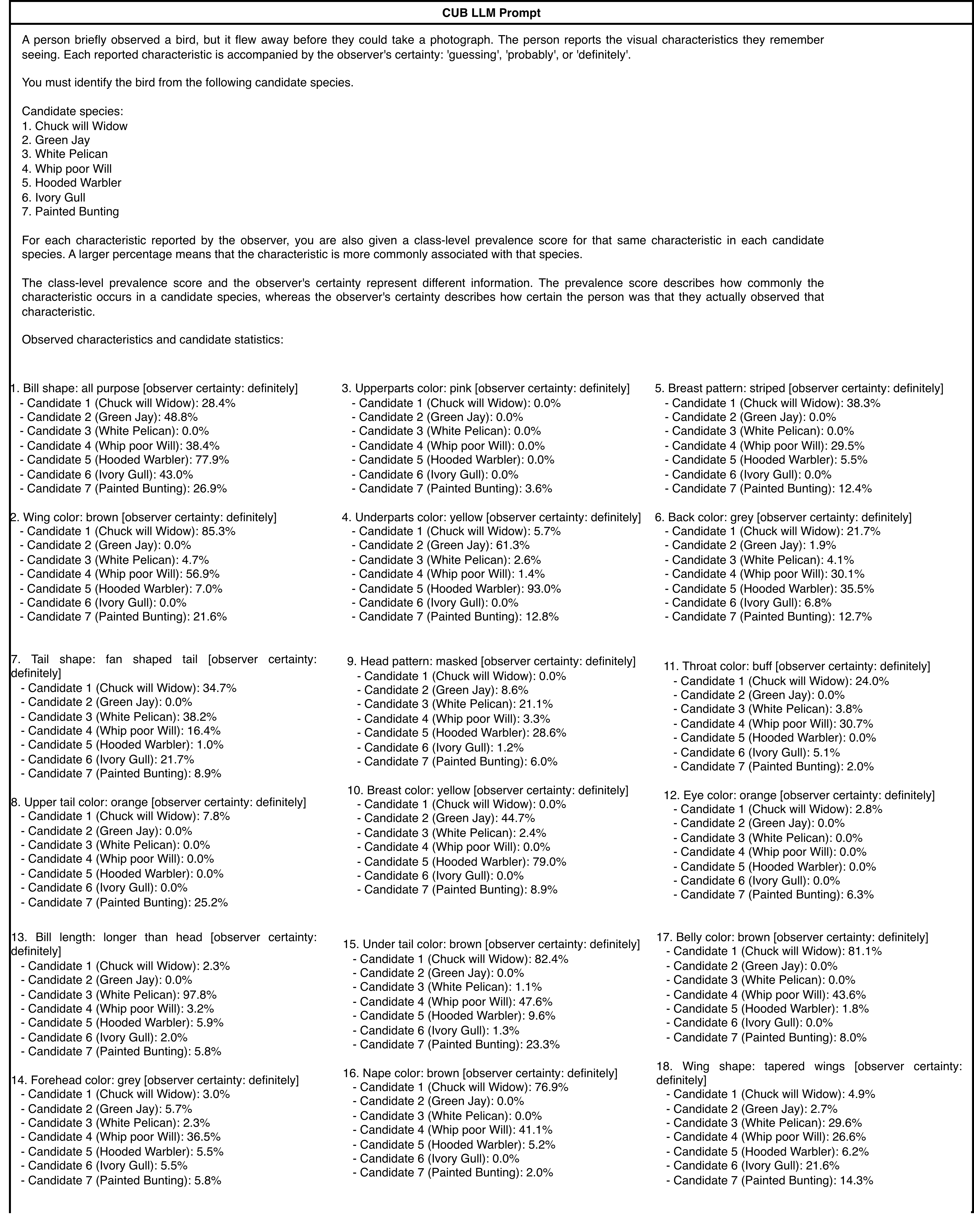}
\end{figure*}
\begin{figure*}[t]
    \centering
    \includegraphics[width=\textwidth, keepaspectratio]{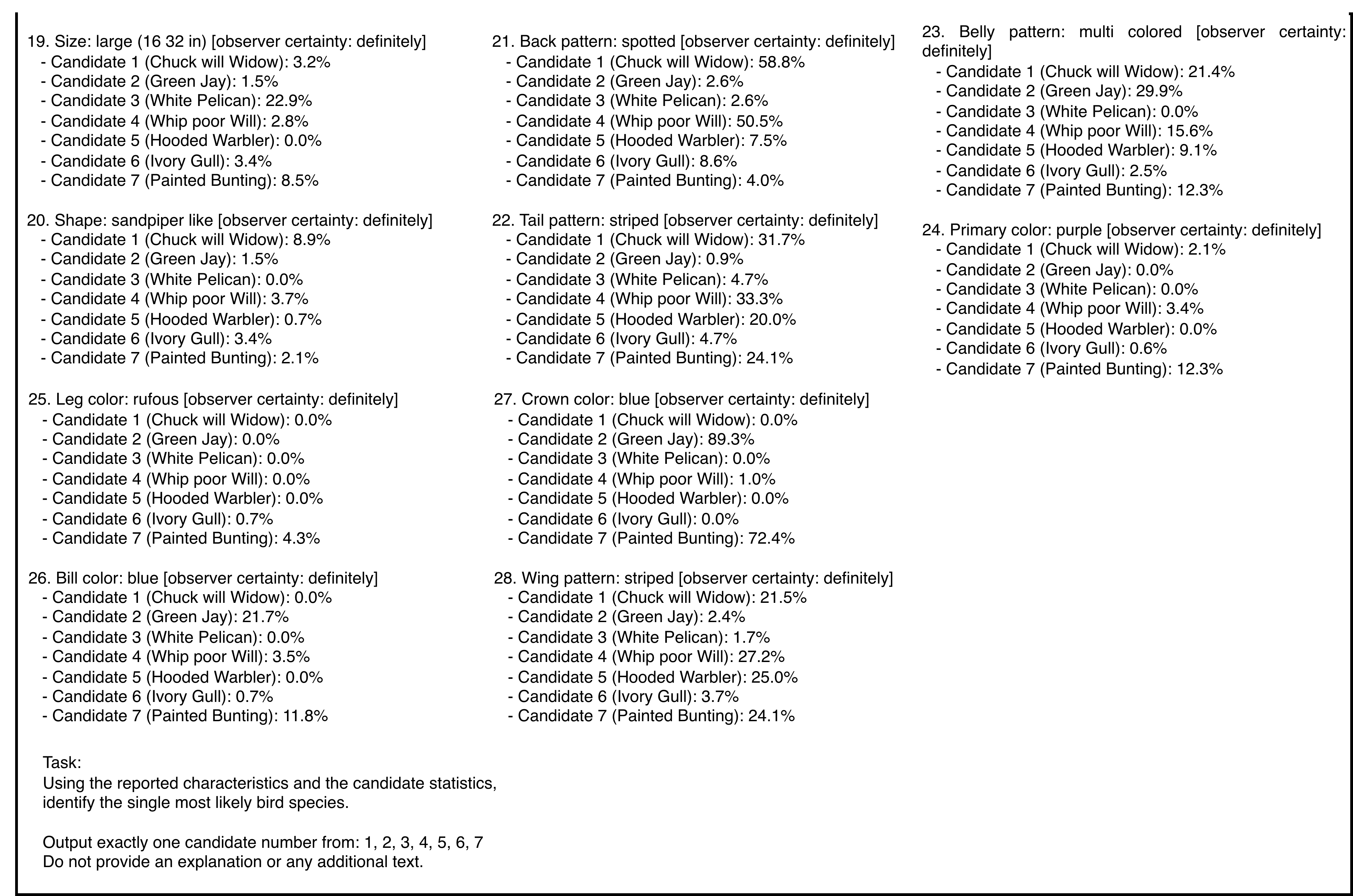}
    \caption{\cub{} LLM Prompt. This is an illustrative example with 7-classes bird species classification task.}
    \label{fig:cub_llm_final_prompt}
\end{figure*}
\subsection{LLM and VLM Prompts}
\begin{figure*}[t]
    \centering
    \includegraphics[width=\textwidth, keepaspectratio]{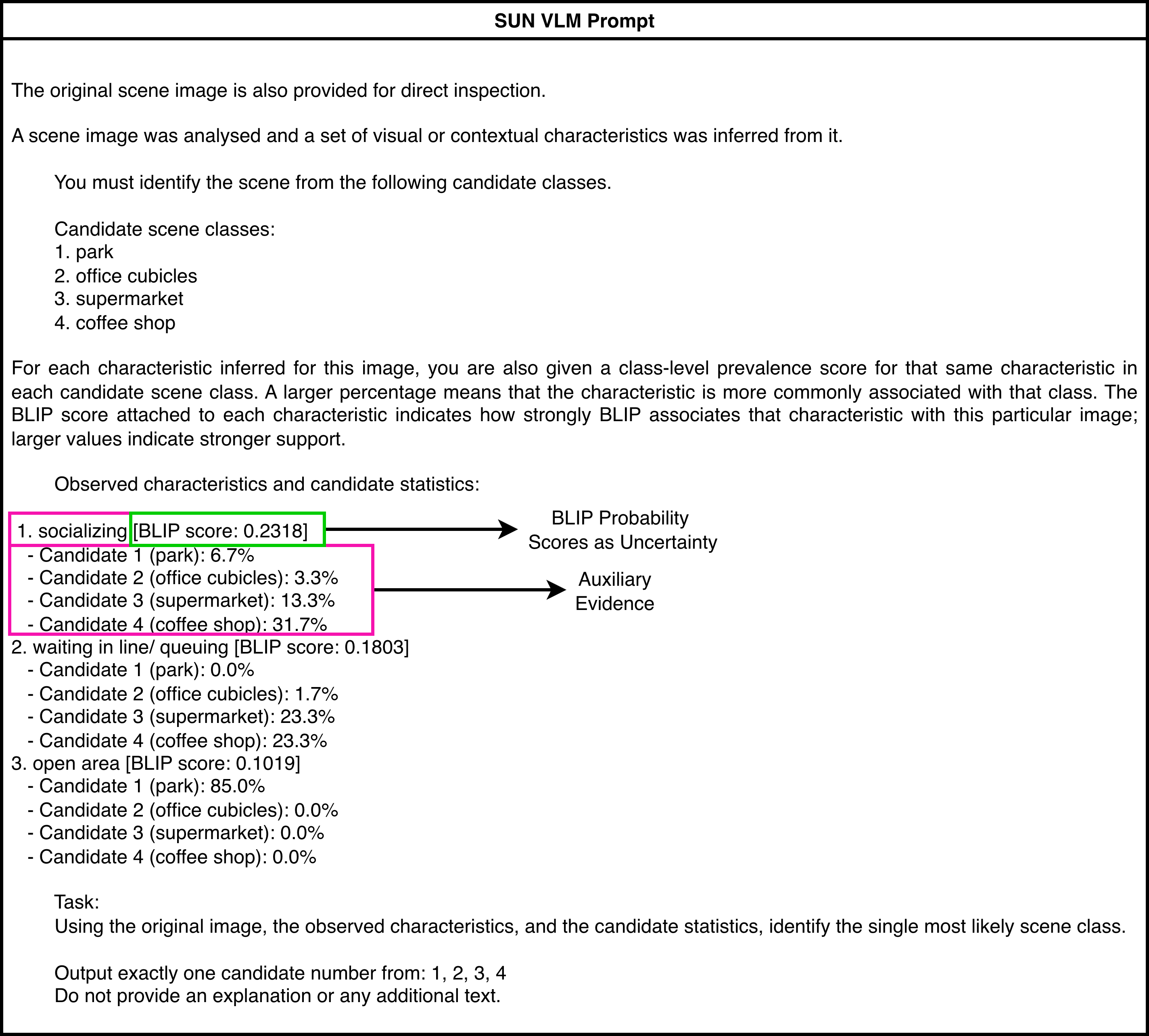}
    \caption{\sun{} scene classification VLM prompt. This is an illustrative example with three auxiliary evidences and their associated uncalibrated BLIP
    image--text matching scores.}
    \label{fig:sun_vlm_prompt}
\end{figure*}

\begin{figure*}[t]
    \centering
    \includegraphics[width=\textwidth, keepaspectratio]{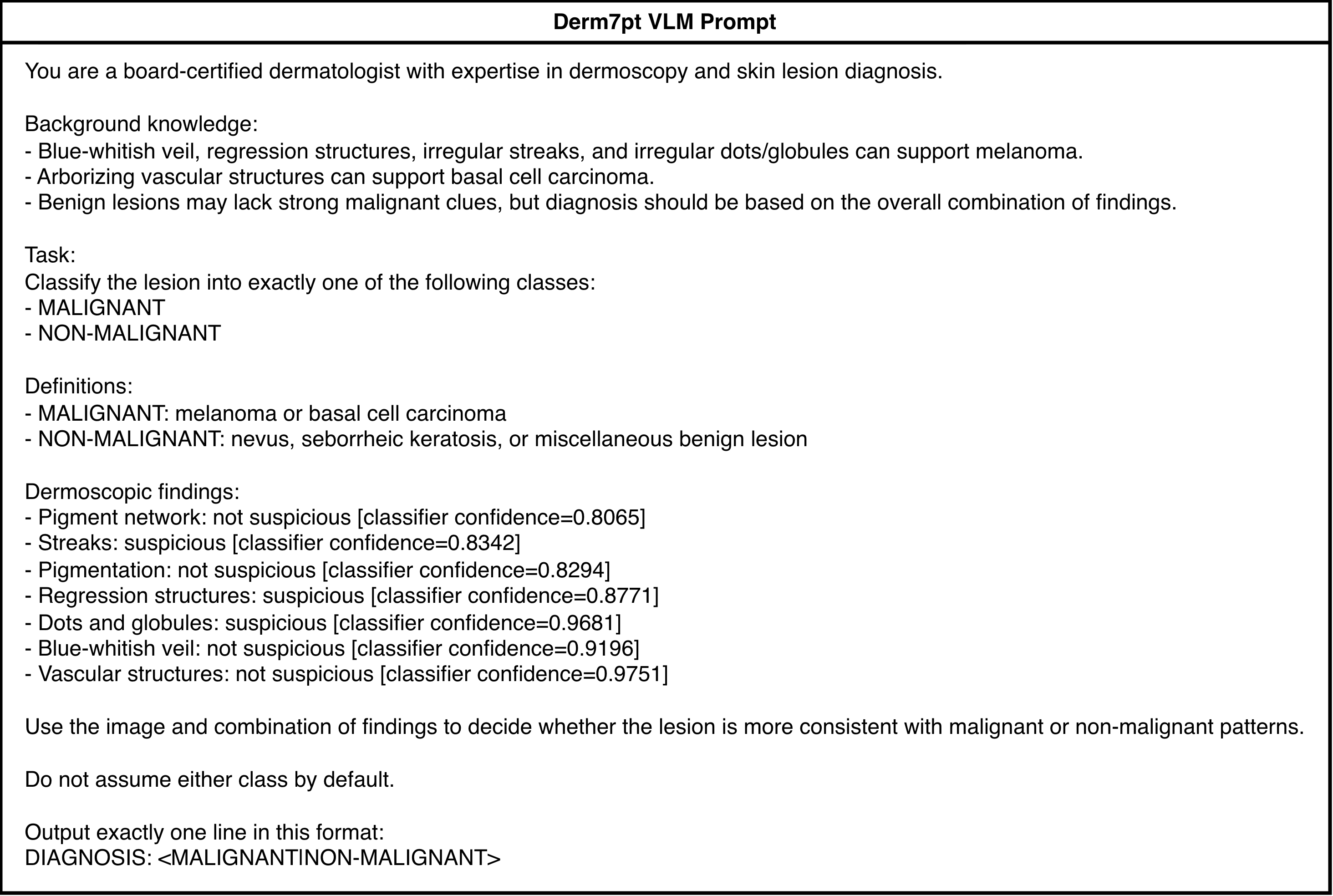}
    \caption{\dermseven{} VLM Prompt. This is an illustrative example with all the seven auxiliary evidence with their associated classifier confidence.}
    \label{fig:derm7pt_vlm_final_prompt}
\end{figure*}

For \sun{}, the LLM prompt closely follows the VLM prompt shown in \figurename~\ref{fig:sun_vlm_prompt}, with the sentence \textit{``The original scene image is also provided for direct inspection.''} removed. For \cub{}, the prompt is likewise largely shared across modalities, following \figurename~\ref{fig:cub_llm_final_prompt}; for the VLM setting, the opening is changed to \textit{``You are given the original bird image.''}. In contrast, \dermseven{} requires modality-specific prompt designs, shown in \figurename~\ref{fig:derm7pt_llm_final_prompt} and \figurename~\ref{fig:derm7pt_vlm_final_prompt} for LLMs and VLMs, respectively. We also observe greater prompt sensitivity on \dermseven{}, which we investigate further in Appendix~\ref{app:derm_prompt_sensitivity}.

\subsection{Prompt Sensitivity: Preliminary Investigation}
\label{app:derm_prompt_sensitivity}

We investigated the sensitivity of Derm7pt classification to the final output instruction. All experiments used the same 395 clinic test cases, all seven ResNet-predicted auxiliary findings, and deterministic inference. The downstream task remained binary classification (non-malignant versus malignant). The dermatologist role, seven-point checklist background knowledge, diagnostic class definitions, and auxiliary evidence were held constant (same as shown in \figurename~\ref{fig:derm7pt_llm_final_prompt}). We evaluated three variants of the final instruction.

\paragraph{P1: Original direct-answer instruction.}

\begin{quote}
\small
\textbf{Instructions:}

Based on the dermoscopic findings, classify the lesion as
NON-MALIGNANT (0) or MALIGNANT (1).

\textbf{Output Format:}

Output exactly one numeric class label: 0 or 1.

Do not provide reasoning, explanations, or any additional text.
\end{quote}

\paragraph{P2: Strict single-character instruction.}

\begin{quote}
\small
\textbf{Instructions:}

Use the dermoscopic findings and the background knowledge to determine whether the lesion is NON-MALIGNANT (0) or MALIGNANT (1).

Perform the classification internally. Do not write out the seven-point checklist, calculate scores in your response, or provide any reasoning.

\textbf{Output Format:}

Your entire response must consist of exactly one character:
0 or 1.

Do not output any other text, punctuation, explanation,
or formatting.
\end{quote}

\paragraph{P3: Intermediate instruction.}

\begin{quote}
\small
\textbf{Instructions:}

Based on the dermoscopic findings, classify the lesion as NON-MALIGNANT (0) or MALIGNANT (1).

\textbf{Output Format:}

Your entire response must consist of exactly one character:
0 or 1.

Do not provide reasoning, explanations, or any additional text.
\end{quote}

\begin{table*}[t]
\centering
\small
\setlength{\tabcolsep}{6pt}
\caption{
Sensitivity of Derm7pt classification to the output instruction formulation. All experiments use the fixed 395-case test set and all seven ResNet-predicted findings ($k=7$). Results are reported in macro-F1. \aux{} provides predicted finding labels only; Full additionally provides both softmax probabilities and entropy; Conf. provides maximum softmax confidence; and Ent. provides predictive entropy.
}
\label{tab:derm_prompt_sensitivity}
\begin{tabular}{llcccc}
\toprule
\textbf{Model} &
\textbf{Prompt} &
\textbf{Aux} &
\textbf{Full} &
\textbf{Conf.} &
\textbf{Ent.} \\
\midrule

\multirow{3}{*}{Qwen3-32B}
& P1 & 0.5053 & 0.6093 & 0.6132 & 0.5807 \\
& P2 & 0.5442 & 0.5775 & 0.5775 & 0.5213 \\
& P3 & 0.5053 & 0.6067 & 0.5963 & 0.5496 \\
\midrule

\multirow{3}{*}{Llama-3.1-8B}
& P1 & 0.6051 & 0.5915 & 0.5772 & 0.5859 \\
& P2 & 0.5083 & 0.5106 & 0.5083 & 0.5111 \\
& P3 & 0.6011 & 0.6194 & 0.6301 & 0.5982 \\
\midrule

\multirow{3}{*}{Phi-4-14B}
& P1 & 0.5442 & 0.5491 & 0.5417$^\dagger$ & 0.5417 \\
& P2 & 0.5513 & 0.5637 & 0.6030 & 0.5564 \\
& P3 & 0.5442 & 0.5467 & 0.5466 & 0.5466 \\

\bottomrule
\end{tabular}

\vspace{2pt}
\parbox{\textwidth}{
\footnotesize
$^\dagger$Phi-4 under P1 with confidence-only evidence achieved
96.96\% parse rate (12 invalid responses). All other reported
configurations achieved 100\% parsing.
}
\end{table*}

Table~\ref{tab:derm_prompt_sensitivity} examines three output-instruction variants while keeping the evidence and domain knowledge fixed. Although the absolute macro-F1 and the magnitude of the uncertainty gain vary across prompts and models, the central trend remains stable, adding uncertainty (in some form of representation) outperforms \aux{} in 8 of the 9 model--prompt combinations. The size of the gain is nevertheless model- and prompt-dependent---for example, \qwenllm{} gains $+10.40$, $+3.33$, and $+10.14$ pp under P1--P3, respectively. This variation motivates treating uncertainty representation together with the downstream model and prompting protocol, rather than assuming that its effect is model-independent.
\begin{wrapfigure}{r}{0.50\textwidth}
\centering
\small
\setlength{\tabcolsep}{3pt}
\captionof{table}{Prompt sensitivity on the 395 held-out Derm7pt clinic images. Macro-F1 is reported for the original VLM prompt and the P3-style prompt adapted from \figurename~\ref{fig:derm7pt_llm_final_prompt}. All evidence based conditions use all seven findings ($k=7$); Image-only uses no auxiliary findings ($k=0$). The final row reports the macro-F1 change from Aux to Aux+Unc in percentage points.}
\label{tab:vlm_prompt_sensitivity}
\vspace{2pt}
\resizebox{\linewidth}{!}{
\begin{tabular}{lcccc}
\toprule
& \multicolumn{2}{c}{\textbf{Qwen2-VL-7B}}
& \multicolumn{2}{c}{\textbf{InternVL3-8B}} \\
\cmidrule(lr){2-3}\cmidrule(lr){4-5}
\textbf{Condition}
& \textbf{Original} & \textbf{P3}
& \textbf{Original} & \textbf{P3} \\
\midrule
\imageonly     & 0.2649 & 0.2270 & 0.4604 & 0.3671 \\
\aux            & 0.5264 & 0.4251 & 0.4739 & 0.5104 \\
\auxunc          & 0.5442 & 0.5183 & 0.4913 & 0.4916 \\
Gold Binary    & 0.6271 & 0.5595 & 0.6377 & 0.6479 \\
Gold Original  & 0.7306 & 0.6982 & 0.7220 & 0.7612 \\
\midrule
$\Delta$(\auxunc, \aux) [pp]
               & +1.8   & +9.3   & +1.7   & -1.9   \\
\bottomrule
\end{tabular}%
}
\end{wrapfigure}


Table~\ref{tab:vlm_prompt_sensitivity} compares our original VLM prompt with a P3-style formulation adapted from the LLM experiments. Under the original prompt, \auxunc{} improves over \aux{} for both \qwenvl{} and \internvl{} ($+1.8$ and $+1.7$ pp, respectively). Under the alternative P3 formulation, the effect becomes more model dependent: \qwenvl{} shows a substantially larger gain of $+9.3$ pp, whereas \internvl{} decreases by $1.9$ pp. Thus, as with the LLMs, prompt formulation changes the magnitude of the uncertainty effect, but the benefit is preserved in three of the four model--prompt combinations. Notably, \goldoriginal{} consistently outperforms \goldbinary{} under both prompts and both VLMs, indicating that richer auxiliary evidence remains useful across prompt formulations. These results further support evaluating uncertainty communication jointly with the downstream model and prompt protocol, rather than assuming a model-independent effect.

\section{A Bernoulli View of Upstream-to-Downstream Uncertainty}
\label{app:unc_to_unc}

For \dermseven{}, each auxiliary predictor solves a binary task and therefore defines a Bernoulli predictive distribution. For sample $i$ and auxiliary finding $j$, let
\[
p_i^j = \Pr(A_i^j=1\mid \mathbf{x}_i),
\qquad
A_i^j \sim \mathrm{Bernoulli}(p_i^j).
\]
The corresponding upstream predictive uncertainty is
\[
h_i^j =
-p_i^j\log p_i^j-(1-p_i^j)\log(1-p_i^j).
\]
The communicated classifier confidence is $u_i^j=\max(p_i^j,1-p_i^j)$; for a binary predictor, $h_i^j$ is therefore a deterministic function of $u_i^j$.

The downstream diagnosis is also binary. Let
\[
Y_i \sim \mathrm{Bernoulli}(q_i),
\qquad
q_i = \Pr(Y_i=1\mid c_i^{\texttt{Unc}}),
\]
where $c_i^{\texttt{Unc}}=(\tau,\mathcal{A},\mathbf{a}_i,\mathbf{u}_i)$ is the uncertainty-aware downstream context. The downstream predictive uncertainty is then
\[
H_i =
-q_i\log q_i-(1-q_i)\log(1-q_i).
\]

This gives the following view of the upstream-to-downstream pathway:
\[
\underbrace{\mathbf{p}_i
\longrightarrow
(\mathbf{a}_i,\mathbf{u}_i)}_{\text{upstream prediction and communicated evidence}}
\longrightarrow
\underbrace{q_i
\longrightarrow
H_i}_{\text{downstream prediction and uncertainty}},
\]
where $\mathbf{p}_i=(p_i^1,\ldots,p_i^k)$.

Importantly, there is no reason to expect a simple mapping $H_i=f(h_i^1,\ldots,h_i^k)$. The downstream model observes not only the uncertainty of each auxiliary predictor but also which findings are predicted, how the findings interact, and the remaining task context. Two samples with identical upstream uncertainty values may therefore induce very different downstream predictive distributions.

\end{document}